\documentclass{amsart}

\usepackage{amsfonts,amsmath,amsthm}
\usepackage{bm}
\usepackage{graphicx}
\usepackage{algorithm}
\usepackage{algorithmic}
\usepackage{multirow}
\usepackage{adjustbox}
\usepackage{pdflscape}

\newtheorem{proposition}{Proposition}

\newtheorem{lemma}{Lemma}

\def\bfI{\mathbf{I}}
\def\bfX{\mathbf{X}}
\def\bfY{\mathbf{Y}}
\def\bfW{\mathbf{W}}
\def\bfK{\mathbf{K}}
\def\bfS{\mathbf{S}}
\def\bfL{\mathbf{L}}
\def\bfZ{\mathbf{Z}}

\def\bfU{\mathbf{U}}
\def\bfM{\mathbf{M}}
\def\bfE{\mathbf{E}}
\def\bfC{\mathbf{C}}
\def\bfR{\mathbf{R}}
\def\bfQ{\mathbf{Q}}
\def\bfP{\mathbf{P}}

\def\bfzero{\mathbf{0}}
\def\bfone{\mathbf{1}}
\def\bfw{\mathbf{w}}
\def\bfx{\mathbf{x}}
\def\bfy{\mathbf{y}}

\def\bfz{\mathbf{z}}
\def\bfc{\mathbf{c}}
\def\bfr{\mathbf{r}}
\def\bfv{\mathbf{v}}

\def\N{\mathcal{N}}
\def\P{\mathcal{P}}
\def\M{\mathcal{M}}

\def\D{\mathcal{D}}

\def\G{\mathcal{G}}
\def\E{\mathcal{E}}
\def\V{\mathcal{V}}
\def\Z{\mathcal{Z}}
\def\S{\mathcal{S}}
\def\T{\mathcal{T}}
\def\A{\mathcal{A}}
\def\W{\mathcal{W}}

\def\bbx{\mathbb{X}}
\def\bby{\mathbb{Y}}
\def\bbE{\mathbb{E}}
\def\bbR{\mathbb{R}}

\def\bbP{\mathbb{P}}
\def\bmeps{\bm{\epsilon}}

\def\Tr{\textrm{Tr}}
\def\KL{\textrm{KL}}
\def\d{\textrm{d}}
\def\st{\textrm{s.t.}}

\title{Probabilistic Dimensionality Reduction via Structure Learning}
\author[L. Wang] {Li Wang}
\thanks{
		L. Wang is with Department of Mathematics, Statistics, and Computer Science, University of Illinois at Chicago, Chicago, USA. E-mail: liwang8@uic.edu }

\begin{document}
\maketitle

\begin{abstract}
We propose a novel probabilistic dimensionality reduction framework that can naturally integrate the generative model and the locality information of data.  
Based on this framework, we present a new model, which is able to learn a smooth skeleton of embedding points in a low-dimensional space from high-dimensional noisy data. The formulation of the new model can be equivalently interpreted as two coupled learning problem, i.e., structure learning and the learning of projection matrix. 
This interpretation motivates the learning of the embedding points that can directly form an explicit graph structure. 
We develop a new method to learn the embedding points that form a spanning tree, which is further extended to obtain a discriminative and compact feature representation for clustering problems.
Unlike traditional clustering methods, we assume that centers of clusters should be close to each other if they are connected in a learned graph, and other cluster centers should be distant. This can greatly facilitate data visualization and scientific discovery in downstream analysis.
Extensive experiments are performed that demonstrate that the proposed framework is able to obtain discriminative feature representations, and correctly recover the intrinsic structures of various real-world datasets.
\end{abstract}

\section{Introduction}

Contemporary simulation and experimental data acquisition technologies enable scientists and engineers to generate progressively large and inherently high-dimensional data sampled from sources with unknown multivariate probability distributions. Data expressed with many degrees of freedom imposes serious problems for data analysis. It is often difficult to directly analyze these datasets in the high-dimensional space, and is desirable to reduce the data dimensionality in order to overcome the curse of dimensionality and associate data with intrinsic structures for data visualization and subsequent scientific discovery. 

Dimensionality reduction is a learning paradigm that transforms high-dimensional data into a low-dimensional representation. Ideally, the reduced representation should correspond to the intrinsic dimensionality of the original data, and it can often be of advantage in practical applications to analyze the intrinsic structure of  data. Examples include clustering of gene expression data and text documents \cite{Ding2004}, and high-dimensional data visualization of image datasets \cite{Roweis2000, Tenenbaum2000}. 
Accordingly, many dimensionality reduction methods have been proposed with the aim to preserve certain information of data. Principal component analysis (PCA) \cite{Jolliffe1986} is a classic method for this purpose. It learns a subspace linearly spanned over some orthonormal bases by minimizing the reconstruction error \cite{Burges2009}. However, a complex structure of the data could be misrepresented by a linear manifold constructed by PCA.

Complex structures have been studied to overcome the misrepresentation issue of PCA. Kernel PCA \cite{Scholkopf1999} first maps the original space to a reproducing kernel Hilbert space (RKHS) by a kernel function and then performs PCA in the RKHS space. Hence, KPCA is a nonlinear generalization of traditional PCA.  
Another approach is manifold learning \cite{Cayton2005}, which aims to find a manifold close to the intrinsic structure of data. By projecting data onto a manifold, the low-dimensional representation of data can be obtained by unfolding the manifold. Isometric feature mapping (Isomap) \cite{Tenenbaum2000} first estimates geodesic distances between data points using shortest-path distances on a $K$-nearest neighbor graph, and then use multidimensional scaling to find points in a low-dimensional Euclidean space where the distance between any two points match the corresponding geodesic distance. Local linear embedding (LLE) \cite{saul2003think} finds a mapping that preserves local geometry where local patches based on $K$-nearest neighbors are nearly linear and overlap with one another to form a manifold. Laplacian eigenmap (LE) \cite{belkin2001laplacian} is proposed based on spectral graph theory, and a $K$-nearest neighbor graph is used to construct a Laplacian matrix. Other methods related to neighborhood graphs are referred to survey papers \cite{Cayton2005,van2009dimensionality}, including maximum variance unfolding (MVU) \cite{weinberger2004learning},  diffusion maps (DM) \cite{lafon2006diffusion}, Hession LLE \cite{donoho2003hessian}, and local tangent space analysis (LTSA) \cite{zhang2004principal}.

Probabilistic models have also been studied for dimensionality reduction. Probabilistic PCA (PPCA) \cite{tipping1999probabilistic} generalizes PCA by applying the latent variable model to the representation of linear relationship between data and its embeddings. Gaussian process latent variable model (GPLVM) \cite{lawrence2005probabilistic} takes an alternative approach to marginalize the linear projection matrix, and then parametrizes covariance matrix using a kernel function. GPLVM with a linear kernel is the dual interpretation of PPCA, while its nonlinear generalization is related to KPCA. Bayesian GPLVM \cite{titsias2010bayesian} maximizes the likelihood of data by marginalizing out both projection matrix and embeddings using variational inference. Maximum entropy unfolding (MEU) \cite{lawrence2012unifying} is proposed to directly model the density of observed data by minimizing Kullback-Leibler (KL) divergence under a set of constraints, and embedding points of data are obtained by maximizing the likelihood of the learned density. t-distributed stochastic neighbor embedding (tSNE) \cite{van2008visualizing} employs a heavy-tailed distribution in the low-dimensional space to alleviate both the crowding problem and the optimization problem of SNE \cite{hinton2002stochastic}, which converts the high-dimensional Euclidean distances between data points into conditional probabilities that represent similarities.

Although the above two classes of methods work well under certain conditions, they lack a unified probabilistic framework to learn robust embeddings from noisy data by applying neighborhood graph to capture the locality information. 
Probabilistic models such as PPCA and GPLVM can deal with noisy data, but they are difficult to incorporate the neighborhood manifold, which has been proved to be effective for nonlinear dimensionality reduction. On the other hand, methods based on neighborhood manifold, such as MVU, LE and LLE, either are hard to learn the manifold structure of a smooth skeleton, or cannot be interpreted as probabilistic models for model selection and noise tolerance.  

In addition, neighborhood graphs used in the above methods are generally constructed from data resided in a high-dimensional space if they are unknown. 
Graph structures that are commonly used in graph based clustering and semi-supervised learning are the $K$-nearest neighbor graph and the $\epsilon$-neighborhood graph \cite{belkin2006manifold}. Dramatic influences of these two graphs on clustering techniques have been studied in \cite{Maier2009}. Since the $\epsilon$-neighborhood graph could result in disconnected components or subgraphs in the dataset or even isolated singleton vertices, the $b$-matching method is applied to learn a better $b$-nearest neighbor graph via loopy belief propagation \cite{Jebara2009}. However, it is improper to use a fixed neighborhood size since the curvature of manifold and the density of data points may be different in different regions of the manifold \cite{elhamifar2011sparse}. 
Moreover, most distance-based manifold learning methods suffer from the curse of dimensionality, i.e., 
there is little difference in the distances between different pairs of data points \cite{beyer1999nearest}. 
Hubs are closely related to the nearest neighbors \cite{radovanovic2010hubs}, that is, points that appear in many
more $K$-nearest neighbors lists than other points, effectively making them ``popular'' nearest neighbors. As alluded by \cite{radovanovic2010hubs}, hubs can have a significant effect on dimensionality reduction and clustering, so we should take hubs into account in a way equivalent to the existence of outliers.  Furthermore, if the data is noisy, a precomputed neighborhood graph to approximate the manifold of data is not reliable any more. As a result, it is less reliable to directly construct $K$-nearest neighbor graphs in a high-dimensional space.

To overcome the issues of constructing graphs, structure learning has had a great success in automatically learning explicit structures from data. 
A sparse manifold clustering and embedding (SMCE) \cite{elhamifar2011sparse} is proposed using  $\ell_1$ norm over the edge weights and $\ell_2$ norm over the errors that measure the linear representation of every data point by using its neighborhood information. Similarly, $\ell_1$ graph is learned for image analysis using $\ell_1$ norm over both the edge weights and the errors for enhancing the robustness of the learned graph \cite{Cheng2010}. Instead of learning directed graphs by using the above two methods, an integrated model for learning an undirected graph by imposing the sparsity on a symmetric similarity matrix and a positive semidefinite constraint on the Laplacian matrix is proposed \cite{lake2010discovering}. These are discriminative models, so they lack the ability to model noise of data. In addition to learning a general graph, a simple principal tree learning algorithm (SimplePPT) \cite{Mao2015} aims to learn a spanning tree from data by minimizing data quantization error and the length of the tree. However, SimplePPT generates principal tree in the original space of the input data, so it is not applicable for dimensionality reduction.

To overcome the above issues, we propose a novel probabilistic dimensionality reduction framework that can naturally integrate the generative model and the locality information of data. 
This framework is formulated in terms of empirical Bayesian inference where the likelihood function models the data generation process using noise model and the loss function penalizes the violations of expected pairwise distances between two original data points and their corresponding embedded points in a neighborhood. 
This framework generalizes the dimensionality reduction approach proposed in our preliminary work \cite{mao2015dimensionality}, which learns a mapping function that transforms data points in a high-dimensional space to latent points in a low-dimensional space such that these latent points directly forms a graph. Our previous work has proved that the proposed method is able to correctly discover the intrinsic structures of various real-world datasets including curves, hierarchies and a cancer progression path. Extensive experiments are conducted to validate our proposed methods for the visualization of learned embeddings for general structures and classification performance. 
The main contributions of this paper are summarized as follows:

1) We propose a novel probabilistic framework for dimensionality reduction, which not only takes the noise of data into account, but also utilizes the neighborhood graph as the locality information. To the best of our knowledge, there is no prior work that can model data generation error and pairwise distance constraints in a unified framework for dimensionality reduction.

2) We present a new model under the proposed framework using $\ell_2$ loss function over the expected distances. Given a neighborhood graph, this model is able to learn a smooth skeleton structure of embedded points and retain the inherent structure from noisy data by imposing the shrinkage of the pairwise distances between data points.

3) To learn an explicit graph structure, we break down the optimization problem of the new model into two components: structure learning and projection matrix learning. By replacing the component of structure learning with the problem of learning an explicit graph, the new model reduces to our preliminary work \cite{mao2015dimensionality}.

4) The connections between the proposed models and various existing methods are discussed including reversed graph embedding, MEU, and the various structure learning methods mentioned above.

The rest of the paper is organized as follows. We first briefly introduce several existing methods from deterministic and probabilistic points of view  in Section \ref{sec:related}. In Section \ref{sec:str-dr}, we propose a unified probabilistic framework for dimensionality reduction and a new model for learning a smooth skeleton from noisy, high-dimensional data. We further generalize this framework for learning an explicit graph structure in Section \ref{sec:explicit-graph} and discuss the connections to various existing works in Section \ref{sec:connect}. 
Extensive experiments are conducted in Section \ref{sec:experiments}. 
We conclude this work in Section \ref{sec:conclusion}. Moreover, proofs and more experimental results are given in the supplementary materials.

\section{Related Work} \label{sec:related}

Let $\bby=\{\bfy_i\}_{i=1}^N$ be a set of $N$ data points where $\bfy_i \in \bbR^D$. The goal of dimensionality reduction is to find a set of embedded data points $ \bbx = \{ \bfx_i \}_{i=1}^N$, where $\bfx_i \in \bbR^{d}$ and $d < D$, satisfying certain assumptions. Next, we briefly introduce several existing dimensionality reduction methods from both deterministic and probabilistic perspectives.

\subsection{Deterministic Methods}
The classic deterministic method for dimensionality reduction is MVU \cite{weinberger2004learning}. Its objective is to maximize the variance of the embedded points subject to constraints such that distances between nearby inputs are preserved. 
MVU consists of three steps. The first step is to compute the $K$-nearest neighbors $\N_i$ of data point $\bfy_i, \forall i$. The second step is to solve the following optimization problem

\begin{small} \vspace{-0.15in}
	\begin{align}
	\max_{\bbx} &~ \sum_{i=1}^N ||\bfx_i||^2 \label{op:mvu}\\ 
	\textrm{s.t.} &~ ||\bfx_i - \bfx_j||^2 = ||\bfy_i - \bfy_j||^2, \forall i, j \in \N_i, \label{con:preserve} \\
	&~ \sum_{i=1}^N \bfx_i = 0, \label{con:origin}
	\end{align}
\end{small}\noindent
where constraints (\ref{con:preserve}) preserve distances between $K$-nearest neighbors and constraint (\ref{con:origin}) eliminates the translational degree of freedom on the embedded data points by constraining them to be centered at the origin. 
Instead of optimizing over $\bbx$, MVU reformulates (\ref{op:mvu}) as a semidefinite programming by learning a kernel matrix $\bfK$ with the $(i,j)$th element denoted by $\kappa_{i,j} = \langle \bfx_i, \bfx_j\rangle_{\mathcal{H}}$ with a semidefinite constraint $\bfK \succeq 0$ for a valid kernel \cite{scholkopf2001learning} where the corresponding mapping function lies in a RKHS $\mathcal{H}$. 
Define $\phi_{i,j} = ||\bfy_i - \bfy_j||^2$ and $\zeta_{i,j} = ||\bfx_i - \bfx_j||^2 = \kappa_{i,i} + \kappa_{j,j} - 2 \kappa_{i,j}$.
The resulting semidefinite programming is 

\begin{small}\vspace{-0.1in}
	\begin{align*}
	\max_{\bfK} \textrm{Tr}(\bfK): \textrm{s.t.}  \sum_{i,j} \kappa_{i,j} = 0, \bfK \succeq 0, \zeta_{i,j} = \phi_{i,j}, \forall i, j \in \N_i,
	\end{align*}
\end{small}\noindent
where $ \langle \sum_{i=1}^N \bfx_i, \sum_{j=1}^N \bfx_j \rangle_{\mathcal{H}} = \sum_{i,j} \kappa_{i,j} =  0$  is a relaxation of (\ref{con:origin}) for ease of kernelization. The last step is to obtain the embedding $\bbx$ by applying KPCA on the optimal $\bfK$.
The distance/similarity information on a neighborhood graph is widely used in the manifold-based dimensionality reduction methods such as locally linear embedding (LLE) and its variants \cite{saul2003think}, and Laplacian Eigenmap (LE) \cite{belkin2001laplacian}.

A duality view of MVU problem has been studied in \cite{xiao2006duality}. Define $\bfE^{i,j}$ as an $N \times N$ matrix consisting of only four nonzero elements: $\bfE^{i,j}[i,i] = \bfE^{i,j}[j,j]=1, \bfE^{i,j}[i,j]=\bfE^{i,j}[j,i]=-1$. The preserving constraints can be rewritten as $\textrm{Tr} ( \bfK \bfE^{i,j} ) = \phi_{i,j}, \forall i, j \in \N_i$. Thus, the dual problem of the above semidefinite programming is given by

\begin{small}\vspace{-0.1in}
	\begin{align}
	\min_{ \{w_{i,j}\} } \sum_{i, j\in\N_i} w_{i,j} \phi_{i,j} : \textrm{s.t.}~ \lambda_{N-1} (\bfL) \geq 1, \bfL = \sum_{i, j\in\N_i} w_{i,j} \bfE^{i,j}, \label{op:dual-mvu}
	\end{align}
\end{small}\noindent
where $w_{i,j}$ is the dual variable subject to the preserving constraint associated to edge $(i,j)$, and $\lambda_{N-1}$ denotes the second smallest eigenvalue of a symmetric matrix \cite{xiao2006duality}..

\subsection{Probabilistic Models}
Probabilistic models are able to take the noise model of data generation into consideration.  The observed data $\bby$ and the embedding $\bbx$ are treated as random variables.
For dimensionality reduction, we associate a set of latent variables $\bfX \in \bbR^{N \times d}$ to a set of observed variables $\bfY \in \bbR^{N \times D}$. 

Latent variable models for dimensionality reduction  generally assume the linear relationship between $\bfx_i$ and $\bfy_i$ with noise given by

\begin{small}\vspace{-0.15in}
	\begin{align}
	\bfy_i = \bfW \bfx_i + \bmeps_i, \forall i, \label{eq:yx}
	\end{align}
\end{small}\noindent
where $\bfW \in \bbR^{D \times d}$ is a linear projection matrix, and $\bmeps_i \in \bbR^D$ is the vector of noise values. Noise is independently sampled from a spherical Gaussian distribution with mean zero and covariance $\gamma^{-1} \bfI_D$ where $\gamma >0$ and $\bfI_D$ is a $D \times D$ identity matrix. Thus, the likelihood of data point $\bfx_i$ is 

\begin{small}\vspace{-0.15in}
	\begin{align}
	p(\bfy_i | \bfx_i, \bfW, \gamma) = \N( \bfy_i | \bfW \bfx_i, \gamma^{-1} \bfI_D ),
	\end{align}
\end{small}\noindent
and the likelihood of the whole data is $p(\bfY | \bfX, \bfW, \gamma) = \prod_{i=1}^N p(\bfy_i | \bfx_i, \bfW, \gamma)$ due to the independently and identically distributed (i.i.d.) assumption of data.

PPCA \cite{tipping1999probabilistic} further assumes that the latent variables $\{ \bfx_i \}_{i=1}^N$ follow a unit covariance zero mean Gaussian distribution

\begin{small}\vspace{-0.15in}
	\begin{align}
	\pi(\bfx_i) = \N(\bfx_i | \bfzero_d, \bfI_d).
	\end{align}
\end{small}\noindent
The projection matrix $\bfW$ is obtained by maximizing the likelihood of the given data

\begin{small}\vspace{-0.15in}
	\begin{align}
	\max_{\bfW} p(\bfY | \bfW, \gamma) \label{op:mll}
	\end{align}
\end{small}\noindent
where $p(\bfY | \bfW, \gamma) = \prod_{i=1}^N p(\bfy_i | \bfW, \gamma)$ and the marginal likelihood of each data is obtained by marginalizing out the latent variable $\bfX$ given by

\begin{small}\vspace{-0.15in}
	\begin{align}
	p(\bfy_i | \bfW, \gamma) &= \int p(\bfy_i | \bfx_i, \bfW, \gamma)  \pi(\bfx_i) d \bfx_i \nonumber\\
	&=\N(\bfy_i | \bfzero, \bfW \bfW^T + \gamma^{-1} \bfI_D ).
	\end{align}
\end{small}\noindent
Tipping and Bishop \cite{tipping1999probabilistic} showed that the principal subspace of the data is the optimal solution of problem (\ref{op:mll}) if $\gamma$ approaches to infinity. Therefore, this model is viewed as a probabilistic version of PCA.

GPLVM \cite{lawrence2005probabilistic} takes an alternative way to obtain marginal likelihood of data by marginalizing out $\bfW$ and optimizing with respect to $\bfX$. Assume the prior distribution of $\bfW$ as 

\begin{small}\vspace{-0.15in}
	\begin{align}
	\pi(\bfW) = \prod_{j=1}^D \N(\bfw_j | \bfzero_d, \bfI_d ).
	\end{align}
\end{small}\noindent
where $\bfw_j$ is the $j$th row of $\bfW$.
The marginal likelihood of the data is obtained by marginalizing out $\bfW$ given by

\begin{small}\vspace{-0.15in}
	\begin{align}
	p(\bfY | \bfX, \gamma) &= \int \prod_{i=1}^N p(\bfy_i | \bfx_i, \bfW, \gamma) \pi(\bfW) d \bfW \nonumber\\
	&= \mathcal{MN}_{N,D}(\mathbf{0}, \widehat{\bfK}, \bfI_D) \label{eq:gplvm}
	\end{align}
\end{small}\noindent
where $\mathcal{MN}_{N,D}$ is the matrix normal distribution with zero mean, sample-based covariance matrix $\widehat{\bfK} = \bfX \bfX^T + \gamma^{-1} \bfI_N$ and feature-based covariance matrix $\bfI_D$ \cite{gupta1999matrix}. GPLVM obtains $\bfX$ by maximizing the marginal likelihood of the data

\begin{small}\vspace{-0.15in}
	\begin{align}
	\max_{\bfX} \log p(\bfY | \bfX, \gamma). 
	\end{align} 
\end{small}\noindent
Lawrence \cite{lawrence2005probabilistic} showed that the optimal solution is equivalent to that obtained by PCA. The merit of this model is that different covariance functions can be incorporated for nonlinear representations since $\widehat{\bfK}$ is of the form of inner product matrix.
Thus, the linear model is called the dual interpretation of PPCA, and the nonlinear model is related to KPCA \cite{Scholkopf1999}.

%

\section{Structured Dimensionality Reduction} \label{sec:str-dr}

We propose a novel dimensionality reduction framework based on regularized empirical Bayesian inference \cite{zhu2014bayesian}, where the unknown embedded data is not only decided by the observed data, but also regulated by manifold structures. Next, we  first present the regularized empirical Bayesian inference, and expectation constraints for capturing manifold structure are then presented. With these two ingredients, we formulate the proposed framework for dimensionality reduction. 

\subsection{Regularized Empirical Bayesian Inference}

Regularized empirical Bayesian inference \cite{zhu2014bayesian} is an optimization formulation of a richer type of posterior inference, by replacing the standard normality constraint with a wide spectrum of knowledge-driven and/or data-driven constraints or regularization. 
Following the notation in \cite{zhu2014bayesian}, let $\M$ denote the space of feasible models, and $\bfM \in \M$ represents an atom in this space. 
We assume that $\Pi$ is absolutely continuous with respect to some background measure $\mu$, so that there exists a density $\pi$ such that $\d \Pi = \pi \d \mu$.  Given a model, let $\D$ be a collection of observed data points, which are assumed to be i.i.d..
Define $\KL( q(\bfM) | \pi(\bfM) ) = \int_{\M} q(\bfM) \log ( q(\bfM) / \pi(\bfM) ) \d \mu(\bfM) $ as the Kullback-Leibler (KL) divergence from $q(\cdot)$ to $\pi(\cdot)$. 

In the presence of unknown parameters, e.g., hyper-parameters, empirical Bayesian inference is necessary  where an estimation procedure such as maximum likelihood estimation is needed. 
Here, we focus on the expectation constraints, of which each one is a function of $q(\bfM)$ through an expectation operator.
For example, let $\bm{\psi} = \{ \psi_1,\ldots,\psi_T \}$ be a set of feature functions, each of which is $\psi_t(\bfM; \D)$ defined on $\M$ and possibly data dependent.
With unknown parameter $\bm{\Theta}$, regularized empirical Bayesian inference is formulated by solving the following optimization problem

\begin{small}\vspace{-0.15in}
	\begin{align}
	\inf_{\bm{\Theta} , q(\bfM)} &~ \KL( q(\bfM) | \pi(\bfM) ) - \int_{\M} \log p(\D | \bfM, \bm{\Theta}) q(\bfM) \d \mu(\bfM) \nonumber\\
	&~ + U( \{ \bbE_{q(\bfM)}[\psi_t(\bfM; \D) ] \}_{t=1}^T  ) \label{op:ebi} \\
	\st &~ q(\bfM) \in \P_{post}, \bm{\Theta} \in \Theta, \nonumber
	\end{align}
\end{small}\noindent
where 
$\bbE_{q(\bfM)}[\psi_t]$ is the expectation of $\psi_t$ over $q(\bfM)$, $U$ is a function of $\{ \bbE_{q(\bfM)}[\psi_t] \}_{t=1}^T $, $\Theta$ is the feasible set of the unknown parameter $\bm{\Theta}$, and $\P_{post}$ is a subspace of distributions.
%
Note that minimizing the first two terms of (\ref{op:ebi}) with respect to $q(\bfM)$ and $\bm{\Theta}$ leads to an optimal solution $\bm{\Theta}^*$, which is equivalent to maximum likelihood estimation, 

\begin{small}\vspace{-0.15in}
	\begin{align}
	\bm{\Theta}^* = \arg \max_{\bm{\Theta} \in \Theta} \log p(\D | \bm{\Theta})
	\end{align}
\end{small}\noindent
where $q(\bfM) = p(\bfM | \D, \bm{\Theta})$. Hence, problem (\ref{op:ebi}) is called regularized empirical Bayesian inference where the regularization term is useful to capture domain knowledge or structure information of data.

\subsection{Expectation Constraints over Pairwise Distances} \label{sec:expect-distance}

Expectation constraints are widely employed for classification problems in the generalized maximum entropy model \cite{Dudik2007} and regularized Bayesian inference model \cite{zhu2014bayesian} . Here, we are particularly interested in defining expectation constraints for dimensionality reduction. 

Given a probabilistic density function $q(\bfM)$, the definition of feature function over data is one of the necessary ingredients to form an expectation constraint. For classification, the feature-label pair of one instance is naturally treated as $\psi_t$. However, this is not suitable for dimensionality reduction, where features from a single instance are not enough to determine the embedding of the whole data. As discussed before, most discriminant methods take pairwise distance as the key information provided by the data. For example, MVU \cite{weinberger2004learning} takes pairwise distances over a neighborhood graph, and LE \cite{belkin2001laplacian} transforms pairwise distances over a neighborhood graph to similarity. Thus, pairwise distance can be reasonably considered as the factor of the feature function for dimensionality reduction. Specifically, in this paper, the feature function $\psi_{i,j}$ represents the difference between pairwise distance of embedding points $\bfx_i$ and $\bfx_j$ and their corresponding distance $\phi_{i,j}$, i.e., $\psi_{i,j}(\bfX; \bfY) = ||\bfx_i - \bfx_j||^2 - \phi_{i,j}$. 

Another necessary ingredient is to determine function $U$, which has significant influence on density function $q(\bfX)$. By incorporating specific prior information of data, we have various choices. 
One choice is to strictly preserve the pairwise distances over a given neighborhood graph, which corresponds to equality constraints (\ref{con:preserve}) in MVU. To achieve this, we define $U(q(\bfX)) = \sum_{i, j \in \N_i} \mathbb{I}( \bbE_{q(\bfX)} [ \psi_{i,j}(\bfX; \bfY)] = 0) $, where $\mathbb{I}(a)$ is an indicator function that equals to $0$ if the $a=0$ is satisfied; otherwise $\infty$. As a result, the optimal $q(\bfX)$ must satisfy $ \bbE_{q(\bfX)} [ ||\bfx_i - \bfx_j||^2] = \phi_{i,j}, \forall i, j\in\N_i$.  


As discussed in \cite{Dudik2007}, we can formulate different expectation constraints over pairwise distances. In this paper, we are more interested in the shrinkage effect of pairwise distances of data to form a smooth skeleton structure in the embedding space.
This can be achieved by defining function $U(\bm{\xi}) = \sum_{i, j\in \N_i} \xi_{i,j}^2 $ where the closeness tolerance $\xi_{i,j}$ is constrained by $\bbE_{q(\bfX)} [ ||\bfx_i - \bfx_j||^2] - \phi_{i,j} \leq \xi_{i,j}, \forall i, j \in \N_i$. If $\xi_{i,j} \leq 0$, we have $\bbE_{q(\bfX)} [ ||\bfx_i - \bfx_j||^2] \leq \phi_{i,j} + \xi_{i,j} \leq \phi_{i,j}$ so that $\bbE_{q(\bfX)} [ ||\bfx_i - \bfx_j||^2]$ cannot be bigger than $\phi_{i,j}$. On the other hand, if $\xi_{i,j} >0$, we have probably $\bbE_{q(\bfX)} [ ||\bfx_i - \bfx_j||^2] \geq \phi_{i,j}$ but $\bbE_{q(\bfX)} [ ||\bfx_i - \bfx_j||^2]$ cannot be bigger than $\phi_{i,j} + \xi_{i,j}$. Thus, the above function $U(\bm{\xi}) = \sum_{i, j\in \N_i} \xi_{i,j}^2$ prefers to shrink the pairwise distance of two original points as the pairwise distance of the corresponding two embedding points, but these constraints allow the violation of expansion of pairwise distance no more than $\xi_{i,j}$ if $\xi_{i,j} > 0$.
These constraints are quite different from the expectation constraints used in \cite{Dudik2007,zhu2014bayesian}. Moreover, they allow us to use efficient optimization tools, which will be illustrated in Section \ref{sec:SPL-s}.

\subsection{Structured Projection Learning} \label{sec:SPL-s}
We propose a new model by incorporating the newly defined expectation constraints into the regularized empirical Bayesian inference framework.
Following PPCA, we treat $\bfX$ as a random variable and $\bfW$ as an unknown parameter. Given a neighborhood graph with a set $\N_i$ as the neighbors of the $i$th vertex of the graph. According to the regularized empirical Bayesian inference, we formulate the following optimization problem

\begin{small}\vspace{-0.15in}
	\begin{align}
	\min_{ \bfW}\min_{q(\bfX), \bm{\xi}} & \textrm{KL}(q(\bfX) || \pi(\bfX)) - \int \log p(\bfY | \bfX, \bfW, \gamma) q(\bfX) d \bfX \nonumber\\
	&~+ C ||\bm{\xi}||^2_F \label{op:ebi-W}\\
	\textrm{s.t.} &~ \bbE_{q(\bfX)} \Big[ ||\bfx_i - \bfx_j||^2 \Big] - \phi_{i,j} \leq \xi_{i,j}, \forall i, j \in \N_i \nonumber \\
	&~ q(\bfX) \in \P_{prob} \nonumber\\
	&~ \bfW^T \bfW = \bfI_d \nonumber,
	\end{align}
\end{small}\noindent
where $C>0$ is a regularization parameter, the orthogonal constraint is added for preventing arbitrarily scaling of the variable $\bfW$, and $\P_{prob}$ represents the feasible set of all density functions over $\bfX$.

The following proposition shows that problem (\ref{op:ebi-W}) has interesting property in terms of its partial dual problem. The proof is given in the supplementary materials.
\begin{proposition} \label{prop:ebi-W}
	Problem (\ref{op:ebi-W}) has an analytic solution given by
	
	\begin{small}\vspace{-0.1in}
		\begin{align}	
		q(\bfX) \propto &\pi(\bfX) p(\bfY | \bfX, \bfW, \gamma) \exp\Big( - \sum_{i,j\in \N_i} s_{i,j} ||\bfx_i - \bfx_j||^2 \Big) \label{eq:post}
		\end{align}
	\end{small}\noindent	
	and $\xi_{i,j} = \frac{s_{i,j}}{2 C}$
	where $\bfS$ and $\bfW$ can be obtained by solving the following optimization problem
	
	\begin{small}\vspace{-0.15in}
		\begin{align}
		\max_{ \bfS } \min_{\bfW}& \frac{d}{2} \log \det ( (\gamma+1) \bfI_N + 4 \bfL ) - \sum_{i, j\in \N_i}\!\!\! s_{i,j} \phi_{i,j} \!\!-\!\! \frac{1}{4 C} ||\bfS||_F^2 \nonumber\\
		& - \frac{\gamma^2}{2} \Tr(  \bfW^T \bfY^T ( (\gamma+1) \bfI_N + 4 \bfL)^{-1} \bfY \bfW ) \label{op:ebi-SW} \\
		\emph{\textrm{s.t.}} &~ s_{i,j} = 0, \forall i, j \not \in \N_i \nonumber \\
		&~ s_{i,j} \geq 0, s_{i,j}=s_{j,i}, \forall i, j, \nonumber\\
		&~ \bfW^T \bfW = \bfI_d, \nonumber
		\end{align}
	\end{small}\noindent	
	and $\bfL = diag(\bfS \bfone) - \bfS$ is the Laplacian matrix with $\bfone$ as the column vector of all ones.	
\end{proposition}

According to the property of Laplacian matrix, we have $\mathbf{L} = \textrm{diag} (\mathbf{S} \mathbf{1}) - \mathbf{S} \succeq 0$ if $\mathbf{S} \geq 0$. In other words, $\mathbf{L}$ is guaranteed to be positive semidefinite for any non-negative $\mathbf{S}$.
Define $\mathbf{Q} = \textrm{diag}( \mathbf{S} \bfone ) - \mathbf{S} + \frac{(\gamma+1)}{4} \mathbf{I}_N$ and $\W=\{\bfW \in \bbR^{D \times d} | \bfW^T \bfW = \bfI_d \}$. Due to the inversion of $\bfQ$ and nonconvexity of the objective function, it is challenging to solve problem (\ref{op:ebi-SW}) globally. 
In order to reach a stationary point, we take the projected subgradient ascend method \cite{Boyd2003} to solve problem (\ref{op:ebi-SW}).
First, we denote a function with respect to $\bfS$ as

\begin{small}\vspace{-0.15in}
	\begin{align}
	g(\bfS) = \max_{\bfW \in \W} h(\bfS, \bfW). 
	\end{align}	
\end{small}\noindent
where $h(\bfS, \bfW) = \frac{\gamma^2}{8} \Tr(  \bfW^T \bfY^T \bfQ^{-1} \bfY \bfW )$.
It is worth noting that $h(\bfS, \bfW)$ is convex over $\bfW$ given any $\bfS$ since $\bfQ$ is positive definite. Thus, we can obtain the subgradient of $g(\bfS)$ as the convex hull of union of subdifferentials of active functions at $\bfS$ given by

\begin{small}\vspace{-0.15in}
	\begin{align}
	\partial g(\bfS)\!\! =\!\! \mathbf{Co} \bigcup \{ \partial h(\bfS, \bfW) |  h(\bfS, \bfW) \!\!=\!\!  h(\bfS, \bfW^*), \forall \bfW \in \W\},
	\end{align}	
\end{small}\noindent
where $\bfW^* = \arg \max_{\bfW \in \W} h(\bfS, \bfW)$. In this paper, we take $ \partial h(\bfS, \bfW^*) $ as the ascend direction. 
Given an $\bfS$, maximizing $h(\bfS, \bfW)$ with respect to $\bfW$  is equivalent to the problem of classical PCA where the covariance matrix is $ \bfY^T \bfQ^{-1} \bfY $. Hence, $\bfW^*$ consists of the eigenvectors corresponding to the $d$ largest eigenvalues  of the covariance matrix.
Let 

\begin{small}\vspace{-0.15in}
	\begin{align*}
	f(\bfS) &= \frac{d}{2} \log \det ( \bfQ )\! - \!\!\sum_{i, j\in \N_i} s_{i,j} \phi_{i,j}  - \frac{1}{4 C} ||\bfS||_F^2 - \max_{\bfW \in \W} h(\bfS, \bfW)
	\end{align*}
\end{small}\noindent
Thus, we can compute subgradient $\partial f(\bfS)$ based on the optimal $\bfW^*$.
Considering the symmetric property of $\bfS$, the derivative of the $\log\det$ term with respect to $s_{i,j}$ and a pair of indexes $(i,j) \in \mathcal{I}_{\N}= \{ j < i \wedge j \in \N_i, \forall i \}$ is obtained by

\begin{small}\vspace{-0.15in}
	\begin{align*}
	\frac{\partial \log \det (\bfQ) }{\partial s_{i,j}}
	&= \textrm{Tr}\left( \left[\frac{\partial \log \det (\bfQ) }{\partial{\mathbf{Q}}}\right]^T\!\!\! \frac{\partial [\textrm{diag}( \mathbf{S} \bfone ) - \mathbf{S} + \frac{(\gamma+1)}{4} \mathbf{I}_N]  }{\partial s_{i,j}}  \right) \\
	&=  \textrm{Tr}(  \mathbf{Q}^{-1}\mathbf{A}_{i,j} )
	\end{align*}
\end{small}\noindent
where  the matrix $\mathbf{A}_{i,j}$ can be represented by

\begin{small}\vspace{-0.15in}
	\begin{align*}
	[\mathbf{A}_{i,j}](m,n) = \left\{ \begin{array}{rl}
	1,  & m=n=i ~\textrm{or}~ m=n=j \\
	-1, & m=i \wedge n=j ~\textrm{or}~ m=j \wedge n=i \\
	0, & \textrm{otherwise}.
	\end{array} 
	\right.
	\end{align*}
\end{small}\noindent
As a result, the subgradient of the objective function $f(\bfS)$ can be computed as, $\forall (i, j)\in \mathcal{I}_{\N}$,

\begin{small}\vspace{-0.15in}
	\begin{align}
	\!\!\!\!\partial_{s_{i,j}}\!\! f(\mathbf{s}) \!\!=\!\!  \frac{1}{2}\textrm{Tr}(  \mathbf{Q}^{-T} ( d \bfI_N \!+\! \frac{\gamma^2 }{4}\bfP  ) \mathbf{A}_{i,j} ) \!-\! \frac{1}{C} s_{i,j} \!-\! 2 ||\bfy_i \!-\! \bfy_j||^2  \label{eq:subgradient}
	\end{align}
\end{small}\noindent
where $\bfP = \bfY \bfW^* {\bfW^*}^T \bfY^T \bfQ^{-1}$.
Finally, we can solve (\ref{op:ebi-SW}) by using the projected subgradient ascend method as

\begin{small}\vspace{-0.15in}
	\begin{align*}
	\mathbf{s}^{(t+1)} = \Pi_{\mathbf{s} \geq 0} \left( \mathbf{s}^{(t)} + \alpha_t  \partial_{\mathbf{s}} f(\mathbf{s}^{(t)}) \right),
	\end{align*}
\end{small}\noindent
where $\Pi$ is the projection on the non-negative set, and $\alpha_t$ is the step size in the $t$th iteration. In order to guarantee the convergence, the step size should have the properties: $\sum_{t=1}^\infty \alpha_t^2 < \infty$ and $\sum_{t=1}^\infty \alpha_t= \infty$. In the experiments, we take a typical example as $\alpha_t = 1/t$ \cite{Boyd2003}. 

\begin{algorithm}[!t]
	\caption{Structured Projection Learning (SPL) }
	\label{alg:spl}
	\begin{small}
		\begin{algorithmic} [1]
			\STATE \textbf{Input:} Data $\bfY$, neighbors $\N_i, \forall i$, reduced dimension $d$ \\
			parameters $\gamma$ and $C$
			\STATE $\bfS=\bfzero$
			\REPEAT 
			\STATE $\mathbf{Q} = \textrm{diag}( \mathbf{S} \bfone ) - \mathbf{S} + \frac{(\gamma+1)}{4} \bfI_N$
			\STATE Perform eigendecomposition $\bfY^T \bfQ^{-1} \bfY = \bfU \bm{\Gamma} \bfU^T $ with $\textrm{diag} (\bm{\Gamma})$ sorted in a descending order
			\STATE $\bfW = \bfU(:,1\!:\!d)$
			\STATE $\bfZ = \frac{\gamma+1}{4} \bfQ^{-1} \bfY \bfW$
			
			\STATE Compute subgradient using (\ref{eq:subgradient})
			\STATE $\mathbf{s}^{(t+1)} = \Pi_{\mathbf{s} \geq 0} \left( \mathbf{s}^{(t)} + \frac{1}{t}  \partial_{\mathbf{s}} f(\mathbf{s}^{(t)}) \right)$ 
			
			\UNTIL{Convergence}
			\STATE \textbf{Output:} Embedding $\bfX=\frac{\gamma}{4} \bfQ^{-1}\bfY\bfW$			
		\end{algorithmic}
	\end{small}	
\end{algorithm}

After learning the posterior distribution (\ref{eq:post}) with optimal $\bfS$ and $\bfW$, we obtain the embedded points $\mathbf{X}$ by maximizing the logarithm of (\ref{eq:post}) as the maximum a posteriori estimation, i.e., $\max_{\mathbf{X}} \log q(\bfX) $, which can be rewritten as

\begin{small}\vspace{-0.15in}
	\begin{align}
	\min_{\mathbf{X} } \frac{1}{2}  \textrm{Tr}(  \mathbf{X}^T (4 \mathbf{L} + (\gamma + 1) \mathbf{I}_N ) \mathbf{X} ) + \gamma \Tr( \bfX^T \bfY \bfW ). \label{op:MAP}
	\end{align}
\end{small}\noindent
Problem (\ref{op:MAP}) is a quadratic optimization problem since matrix $4 \mathbf{L} + (\gamma + 1) \mathbf{I}_N$ is positive definite. Thus, we can obtain an analytic solution by setting its derivative with respect to $\bfX$ to zero, given by

\begin{small}\vspace{-0.15in}
	\begin{align}
	\bfX = \gamma (4 \mathbf{L} + (\gamma + 1) \bfI_N )^{-1} \bfY \bfW = \frac{\gamma}{4} \bfQ^{-1}\bfY\bfW.
	\end{align}
\end{small}\noindent

As observed, there is a trivial solution for above objective, i.e., $\mathbf{X}=0$ if $\gamma=0$. To overcome this issues, we have the following observation: the posterior distribution is the matrix normal distribution \cite{gupta1999matrix} given by

\begin{small}\vspace{-0.15in}
	\begin{align}
	q(\bfX) \sim \mathcal{MN}_{N,d} (\mathbf{0}, \Sigma, \mathbf{I}_d )
	\end{align}
\end{small}\noindent
where $\Sigma = (4 \mathbf{L} + \mathbf{I}_N)^{-1}$ is the sample-based covariance matrix and can also be interpreted as a regularized Laplacian kernel with regularization parameter $\lambda > 0$ \cite{smola2003kernels}. As a result, we can apply KPCA on $\Sigma$ to achieve the embedded data points if $\gamma=0$ to avoid trivial solution. The pseudo-code of our proposed structured projection learning (SPL) is described in Algorithm \ref{alg:spl}.

\subsection{Algorithm Analysis} \label{sec:aa}

The computational complexity of Algorithm \ref{alg:spl} can be estimated as follows: solving problem (\ref{op:ebi-SW}) takes approximately $O(N^{2.37})$ for computing logdet and inversion of matrix $\mathbf{Q}$ at each iteration; computing subgradient and function value of $f(\bfS)$  takes $O(N^3)$ due to the eigendecomposition. The time complexity of Algorithm \ref{alg:spl} takes the order of $O(N^3)$. Our method can also leverage the fast eigendecomposition methods for finding a small number of large eigenvalues and eigenvectors. Thus, the computational complexity of Algorithm \ref{alg:spl} is same as that in most of spectral based methods, but is much faster than semidefinite programming used in MVU. The theoretical convergence analysis of Algorithm \ref{alg:spl} follows the projected subgradient method \cite{Boyd2003}.

Algorithm \ref{alg:spl} takes two parameters into account, $C$ and $\gamma$, except the low-dimensional parameter $d$, which is a common parameter for dimensionality reduction and will not be discussed in this paper. Parameter $C$ regulates the error tolerance of pairwise distances between original points and embedding points. The larger the $C$ is, the smaller the error tolerance is imposed. In the case of $C=\infty$, the model does not allow the error.
Parameter $\gamma$ controls the noise of data in the generative model (\ref{eq:yx}). More interestingly, this parameter plays an important role on balancing two distinct models: deterministic model (\ref{op:mvu}) and generative model (\ref{eq:yx}). The role becomes clear by investigating the proposed unified model (\ref{op:ebi-SW}). Specifically, problem (\ref{op:ebi-SW}) only learns the similarity matrix $\bfS$ using pairwise distances of original data as input if $\gamma=0$. On the other hand, if $\gamma > 0$, the random noise of original points is simultaneously incorporated by regulating similarity matrix learning of deterministic model and data reconstruction of generative model. The merit of non-zero $\gamma$ leads to an easy embedding process and meanwhile maintaining the intrinsic structure of data in low-dimensional space, which will be investigated in Section \ref{sec:SPL}.

The proposed embedding framework provides a novel way to automatically learn a sparse positive similarity matrix $\mathbf{W}$ from a set of pairwise distances, and the sparse positive similarity matrix is purposely designed for learning the embedded points. This also provides a probabilistic interpretation why MVU takes KPCA as the embedding method after learning a kernel matrix.

\section{Learning Explicit Graph Structure} \label{sec:explicit-graph}

For certain applications, we know the explicit representation of latent graph structure that generates the observed data, but both the embedded points and the correspondence between vertexes of the graph and the observed data are unknown. In this section, we adapt structured projection learning to learn an explicit graph structure, so that the learned embedded points reside on the optimal graph inside a set of feasible graphs with the given graph representation in the latent space.

\subsection{Explicit Graph Structure Learning}

Before presenting the model for explicitly learning a graph structure, we first introduce an important result.
\begin{proposition} \label{prop:transform}
	%
	Given an $\bfS$, $\min_{\bfW} -h(\bfS, \bfW)$ is equivalent to the following optimization problem
	
	\begin{small}\vspace{-0.15in}
		\begin{align}
		\min_{\bfW \in \W, \bfZ}& \frac{1}{2 (\gamma+1)} ||\bfY - \bfZ \bfW^T ||_F^2 + \frac{2}{(\gamma+1)^2} \Tr( \bfZ^T \bfL \bfZ ) \nonumber\\
		& - \frac{1}{2(\gamma+1)} ||\bfY||_F^2, 
		\end{align}	
	\end{small}	\noindent
	where $\bfZ = \frac{\gamma+1}{4} \bfQ^{-1} \bfY \bfW$.
\end{proposition}
\noindent According to the property of Laplacian matrix and the above proposition, we reformulate the problem (\ref{op:ebi-SW}) as

\begin{small}\vspace{-0.15in}
	\begin{align}
	\max_{\bfS} &~ \min_{\bfW \in \W}  -\frac{d}{2} \log \det ( \bfQ ) + \langle \bfS, \Phi_{\bfY} \rangle + \frac{1}{4 C} ||\bfS||_F^2 \nonumber\\
	& + \frac{\gamma^2}{2 (\gamma+1)} ||\bfY - \bfZ \bfW^T ||_F^2 + \frac{2\gamma^2}{(\gamma+1)^2} \Tr( \bfZ^T \bfL \bfZ ), \label{op:maxmin-linear}
	\end{align}
\end{small}\noindent
where $\bfZ$ is a matrix according to Proposition \ref{prop:transform}, $\Phi_{\bfY}$ is a distance matrix with the $(i,j)$th element as $||\bfy_i - \bfy_j||^2$. From (\ref{op:maxmin-linear}), given $\bfS$, we have

\begin{small}\vspace{-0.15in}
	\begin{align}
	\min_{ \bfW, \bfZ}&~  \frac{1}{2 \gamma} ||\bfY - \bfZ \bfW^T ||_F^2 + \frac{2}{\gamma^2} \Tr( \bfZ^T \bfL \bfZ ) \label{op:graph-criterion}\\
	\textrm{s.t.} &~ \bfW^T \bfW = \bfI_d, \nonumber
	\end{align}
\end{small}\noindent
by  removing the following three terms that regulate $\bfS$, i.e., $\frac{d}{2} \log \det ( \gamma \bfI_N + 4 \bfL ) $,  $\langle \bfS, \Phi_{\bfY} \rangle$, and $\frac{1}{4 C} ||\bfS||_F^2$.
Thus, we can view model (\ref{op:ebi-SW}) as an approach to learn $\bfS$ from data by simultaneously preserving expected distances and optimizing (\ref{op:graph-criterion}) as the learning criterion for dimensionality reduction based on a graph. Similarly, we can also learn an explicit graph structure by incorporating known constraints of certain graph structures and minimizing criterion (\ref{op:graph-criterion}).

Following the above annotations, we can define a general graph representation. Let $\G=(\V,\E)$ be an undirected graph, where $\V=\{ V_1,\ldots, V_N\}$ is a set of vertices and $\E$ is a set of edges. Suppose that every vertex $V_i$ corresponds to a point $\bfz_i \in \Z \subset \mathbb{R}^d$, which lies in an intrinsic space of dimension $d$. Denote the weight of edge $(V_i, V_j)$ as $s_{i,j}$, which represents the similarity (or connection indicator) between $\bfz_i$ and $\bfz_j$ in the intrinsic space $\Z$. We assume that matrix $\bfS \in \S$ with the $(i,j)$th element as $s_{i,j}$ can be used to define the representation of a latent graph, where $\S$ is a set of feasible graphs with the given graph representation. 

By combining the above ingredients, we formulate the following optimization problem, given by

\begin{small}\vspace{-0.15in}
	\begin{align}
	\min_{ \bfS, \bfW, \bfZ}&~  \frac{1}{2 \gamma} ||\bfY - \bfZ \bfW^T ||_F^2 + \frac{2}{\gamma^2} \Tr( \bfZ^T \bfL \bfZ ) \label{op:spl-graph}\\
	\textrm{s.t.} &~ \bfW^T \bfW = \bfI_d, \bfS \in \S. \nonumber
	\end{align}
\end{small}\noindent 
Problem (\ref{op:spl-graph}) is more flexible than problem (\ref{op:graph-criterion}) since the graph structure represented by $\bfS$ in  (\ref{op:spl-graph}) can be directly controlled according to $\S$, but (\ref{op:graph-criterion}) cannot, even though they share the same objective function.
Thus, formulation (\ref{op:spl-graph}) is a general framework for dimensionality reduction by learning an intrinsic graph structure in a low-dimensional space. In order to instantiate a new method, we have to specify the feasible set  $\S$ of graphs. 

\subsection{Dimensionality Reduction via Learning a Tree } \label{sec:drtree-begin}

We investigate a family of tree structures, which can be used to deal with various real world problems.

Given a connected undirected graph $\G=(\V,\E)$ with a cost $c_{i,j}$ associated with edge $(V_i,V_j) \in \E, \forall i, \forall j$, let $\T = (\V, \E_{\T})$ be a tree with the minimum total cost and $\E_{\T}$ be the edges forming the tree.
In order to represent and learn a tree, we consider $\{ s_{i,j} \}$ as binary variables where $s_{i,j}=1$ if $(V_i,V_j) \in \E_{\T}$, and $0$ otherwise. Denote $\bfS=[s_{i,j}] \in \{0,1\}^{N \times N}$.
The integer linear programming formulation of minimum spanning tree (MST) can be written as: 
$\min_{\bfS \in \S_0} \sum_{i,j} s_{i,j} c_{i,j},$
where $\S_0 = \{ \bfS \in \{0,1\}^{N \times N}\} \cap \S'$ and $\S'= \{\bfS = \bfS^T\} \cap \{ \frac{1}{2} \sum_{i,j} s_{i,j} = |\V| -1 \} \cap \{ \frac{1}{2} \sum_{ V_i \in \A, V_j \in \A} s_{i,j} \leq |\A|-1, \forall \A \subseteq \V  \}$. The first constraint of $\S'$ enforce the symmetric connection of undirected graph, e.g. $s_{i,j} = s_{j,i}$.
The second constraint states that the spanning tree only contains $|\V|-1$ edges. The third constraint imposes the acyclicity and connectivity properties of a tree. It is difficult to solve an integer programming problem optimally. Instead, we resort to a relaxed problem by letting $s_{i,j} \geq 0$, that is,

\begin{small}
	\vspace{-0.1in}
	\begin{align}
	\min_{\bfS \in \S_{\T}} &~ \sum_{i,j} s_{i,j} c_{i,j}, \label{op:RelaxLP}
	\end{align}
\end{small}\noindent
where the set of linear constraints over convex domain is given by $\S_{\T} = \left\{ \bfS \geq 0 \right\} \cap \S'$. 
Problem (\ref{op:RelaxLP}) can be solved by Kruskal's algorithm \cite{Cheung2008}.

Let $\lambda = \frac{8}{\gamma}$. We can equivalently rewrite (\ref{op:spl-graph}) as the following optimization problem

\begin{small}
	\vspace{-0.15in}
	\begin{align}
	\min_{ \bfW, \bfZ, \bfS } & \sum_{i=1}^N || \bfy_i - \bfW \bfz_i ||^2 + \frac{\lambda}{2} \sum_{i,j} s_{i,j} || \bfz_i - \bfz_j  ||^2 \label{op:dimred-tree} \\
	\textrm{s.t.} &~ \bfW^T \bfW = \bfI_d, ~~\bfS \in \S, \nonumber 
	\end{align}
\end{small}\noindent
where $\bfW = [\bfw_1,\ldots,\bfw_d] \in \bbR^{D \times d}$ is an orthogonal set of $d$ linear basis vectors $\bfw_l \in \bbR^{D}, \forall l$, $\bfZ = [\bfz_1,\ldots,\bfz_N]^T \in \bbR^{N \times d}$ is represented by the projected data points of $\D$ in the low-dimensional space $\bbR^d$, $\bfS = [s_{i,j}] \in \bbR^{N \times N}$ is an adjacent matrix of a tree $\T=( \V, \E_{\T} )$ where $\E_{\T}= \{(i,j) : s_{i,j} \not =0 \}$.

If $\lambda = 0$, problem (\ref{op:dimred-tree}) is equivalent to the optimization problem of PCA, otherwise the data points $\bby$ are mapped into a low-dimensional space where $\{\bfz_i\}_{i=1}^N$ form a tree. Therefore, PCA is a special case of problem (\ref{op:dimred-tree}). Another important observation is the distances between any two latent points are computed in a low-dimensional space, i.e., $ || \bfz_i - \bfz_j  ||^2$ since latent points $\{\bfz_i\}_{i=1}^N$ are in $\bbR^d$. As a result, problem (\ref{op:dimred-tree}) can effectively mitigate the curse of dimensionality. For ease of reference, we name the problem (\ref{op:dimred-tree}) with $\lambda >0$ as dimensionality reduction tree (DRTree).

\subsection{Discriminative DRTree}

DRTree projects data points in a high-dimensional space to latent points that directly form a tree structure in the low-dimensional space. However, the tree structure achieved might be at the risk of losing clustering information. In other words, some data points are supposed to form a cluster, but they are scattered to different branches of the tree, and distances between them on the intrinsic structure become large. 

To incorporate the discriminative information, we introduce another set of latent points $\{\bfc_k\}_{k=1}^K$ as the centers of $\{ \bfz_i \}_{i=1}^N$ where $\bfc_k \in \bbR^d$ so as to minimize the trade-off between the objective functions of $K$-means and DRtree. As a result, we formulate the following optimization problem

\begin{small} \vspace{-0.13in}
	\begin{align}
	\min_{\bfW, \bfZ, \bfS, \bfC,\bbP} &~ \sum_{i=1}^N || \bfy_i - \bfW \bfz_i ||^2 + \frac{\lambda}{2} \sum_{ k,k'} s_{k,k'} || \bfc_{k} - \bfc_{k'} ||^2 \nonumber \\
	&~ + \gamma \sum_{k=1}^K \sum_{j \in \P_k}|| \bfz_j - \bfc_k ||^2 \label{op:dimred-kmeans-tree}\\
	\textrm{s.t.} &~ \bfW^T \bfW = \bfI_d,~~\bfS \in \S_{\T}, \nonumber
	\end{align}
\end{small}\noindent
where the third term of the objective function is same as the objective function of $K$-means, $\bbP = \{ \P_1, \ldots, \P_K \}$ is a partition of $\{1, \ldots, N\}$, $\bfC = [\bfc_1,\ldots,\bfc_K]^T \in \bbR^{K \times d}$ and $\gamma \geq 0$ is a trade-off parameter between the objective function of DRTree and empirical quantization error of latent points $\{\bfz_i\}_{i=1}^N$ and $\{\bfc_k\}_{k=1}^K$.

Unlike problem (\ref{op:dimred-tree}),  problem (\ref{op:dimred-kmeans-tree}) is now regularized on centers $\{\bfy_k\}_{k=1}^K$ instead of $\{\bfz_i\}_{i=1}^N$. However, problem (\ref{op:dimred-tree}) is a special case of problem (\ref{op:dimred-kmeans-tree}) if $K = N$ and $\gamma \rightarrow \infty$ since the third term can be removed without changing the optimal solution of (\ref{op:dimred-kmeans-tree}) due to $\bfz_i = \bfy_i, \forall i$ at optimum. In this case, problems (\ref{op:dimred-tree}) and (\ref{op:dimred-kmeans-tree}) are equivalent. Except for the special case, problem (\ref{op:dimred-kmeans-tree}) is able to achieve discriminative and compact feature representation for dimensionality reduction since clustering objective and DRTree are optimized in a unified framework.

The hard partition imposed by $K$-means, however, has several drawbacks. First, parameter $K$ is data-dependent, so it is hard to set properly. 
Second, it is sensitive to noise, outliers, or some data points that cannot be thought of as belonging to a single cluster \cite{Filippone2008}. Soft partition methods such as Gaussian mixture modeling have also been used in modeling principal curves \cite{Bishop1998, Tibshirani1992}. However, the likelihood of a Gaussian mixture model tends to be infinite when a singleton is formed \cite{Tibshirani1992}. 
To alleviate the problems from which the aforementioned methods suffer, we propose to replace the hard partition $K$-means with a relaxed regularized empirical quantization error given by

\begin{small} \vspace{-0.13in}
	\begin{align}
	\min_{\bfW, \bfZ, \bfS, \bfC,\bfR} & \sum_{i=1}^N || \bfy_i - \bfW \bfz_i ||^2 + \frac{\lambda}{2} \sum_{ k,k' } s_{k,k'} || \bfc_{k} - \bfc_{k'} ||^2 \nonumber \\
	& + \gamma \left [\sum_{k=1}^K \sum_{i=1}^N r_{i,k} || \bfz_i - \bfc_k ||^2 + \sigma \Omega(\bfR) \right] \label{op:dimred-kmeans-relax-tree}\\
	\textrm{s.t.} &~ \bfW^T \bfW = \bfI_d,~~\bfS \in \S_{\T},~ \sum_{k=1}^K r_{i,k} = 1, r_{i,k} \geq 0, \forall i, \forall k, \nonumber
	\end{align}
\end{small}\noindent
where $\bfR \in \bbR^{N \times K}$ with the $(i,k)$th entry as $r_{i,k}$, $\Omega(\bfR) = \sum_{i=1}^N \sum_{k=1}^K r_{i,k} \log r_{i,k}$ is the negative entropy regularization, and $\sigma >0$ is the regularization parameter. The negative entropy regularization transforms hard assignment used in $K$-means to soft assignment used in Gaussian mixture models \cite{Mao2015}, and is also used in other tasks \cite{Mao2015-tnnls}.

\begin{algorithm}[!t]
	\caption{Discriminative DRTree (DDRTree) }
	\label{algo2}
	\begin{small}
		\begin{algorithmic} [1]
			\STATE \textbf{Input:} Data matrix $\bfY$, parameters $\lambda$, $\sigma$ and $\gamma$
			\STATE Initialize $\bfZ$ by PCA
			\STATE $K=N$, $\bfC = \bfZ$ 
			\REPEAT 
			\STATE  $c_{k.k'} = || \bfc_k - \bfc_{k'}||^2, \forall k, \forall k'$
			\STATE  Obtain $\bfS$ by solving (\ref{op:RelaxLP}) via Kruskal's algorithm
			\STATE  $\bfL = \textrm{diag}(\bfS \bfone) - \bfS$
			\STATE  Compute $\bfR$ with each element as (\ref{op:2-R})
			\STATE  $\Gamma = \textrm{diag}(\bfone^T \bfR)$
			\STATE  $\bfQ =  \frac{1}{1+\gamma} \left[ \bfI + \bfR \left( \frac{1+\gamma}{\gamma} \left(\frac{\lambda}{\gamma} \bfL + \Gamma\right) - \bfR^T \bfR \right)^{-1} \bfR^T  \right] $
			\STATE   Perform eigendecomposition \\
			$\bfY^T \bfQ \bfY = \bfU \Lambda \bfU^T$ \\
			and $\textrm{diag}(\Lambda)$ is sorted in a descending order.
			\STATE  $\bfW = \bfU(:,1:d)$
			\STATE $\bfZ = \bfQ \bfY \bfW$
			\STATE  $\bfC = \left(\frac{\lambda}{\gamma} \bfL + \Gamma \right)^{-1} \bfR^T \bfZ$
			\UNTIL{Convergence}
		\end{algorithmic}
	\end{small}
\end{algorithm}

The following proposition shows that problem (\ref{op:dimred-kmeans-relax-tree}) with respect to $\{\bfC, \bfR\}$ by fixing the remaining variables is equivalent to the mean shift clustering method \cite{Cheng1995}, which is able to determine the number of clusters automatically and initialize centers $\{\bfc_k\}_{k=1}^K$ by latent points $\{\bfz_i\}_{i=1}^N, \forall i$ if $K=N$. 
The following lemma further shows that the optimal solution $\bfR$ has an analytical expression if $\bfC$ is given.
\begin{lemma} \label{prop:R}
	Given $\{\bfW, \bfZ, \bfS, \bfC\}$, Problem (\ref{op:dimred-kmeans-relax-tree}) has the optimal solution $\bfR$ given by the following analytical form, $\forall k, \forall i$
	
	\begin{small}\vspace{-0.16in}
		\begin{align}
		r_{i,k} = { \exp\left( -{ || \bfz_i - \bfc_k ||^2 }/{\sigma} \right) } \Big/ { \sum_{k=1}^K \exp \left( -{ || \bfz_i - \bfc_k||^2 }/{\sigma} \right) }. \label{op:2-R}
		\end{align}
	\end{small}\noindent
\end{lemma}

\begin{proposition} \label{prop:mean-shift}
	Given $\{ \bfW, \bfZ, \bfS \}$, $\lambda = 0$ and assuming $K=N$, problem (\ref{op:dimred-kmeans-relax-tree}) with respect to $\{\bfC, \bfR \}$ can be solved by a mean shift clustering method by initializing  $\bfC=\bfZ$.
\end{proposition}

The key difference between problem (\ref{op:dimred-kmeans-relax-tree}) and the traditional mean shift is that the latent points $\{\bfz_i\}_{i=1}^N$ in our model are variables and can be affected by dimensionality reduction and tree structure learning. We also build a connection between problem (\ref{op:dimred-kmeans-relax-tree}) and problem (\ref{op:dimred-kmeans-tree}) as shown in the following proposition.

\begin{proposition} \label{prop:k-means}
	If $\sigma \rightarrow 0$, (\ref{op:dimred-kmeans-relax-tree}) is equivalent to (\ref{op:dimred-kmeans-tree}).
\end{proposition}

The above properties of problem (\ref{op:dimred-kmeans-relax-tree}) facilitates the setting of parameters in different contexts of applications. In the case of dimensionality reduction, discriminative information might be important for some applications such as clustering problems, and Proposition \ref{prop:mean-shift} provides a natural way to form a cluster without predefining the number of clusters. In the case of finding $K$ clusters, we prefer problem (\ref{op:dimred-kmeans-tree}) to (\ref{op:dimred-kmeans-relax-tree}) since (\ref{op:dimred-kmeans-tree}) is formulated in terms of K clusters directly. According to Proposition \ref{prop:k-means}, the purpose of clustering can also be achieved by solving problem (\ref{op:dimred-kmeans-relax-tree}) with a small $\sigma$. 

Alternating structure optimization \cite{Ando2005} is  used to solve problem (\ref{op:dimred-kmeans-relax-tree}). We first partition variables into two groups $\{ \bfW, \bfZ, \bfC \}$ and $\{\bfS, \bfR\}$, and then solve each subproblem iteratively until the convergence is achieved. 

Given $\{\bfS, \bfR\}$, we can obtain an analytical solution by solving problem (\ref{op:dimred-kmeans-relax-tree}) with respect to $\{ \bfW, \bfZ, \bfC\}$, which is discussed in Proposition \ref{prop:ddr-WZY}. Before presenting Proposition \ref{prop:ddr-WZY}, we first state a necessary condition of the proposition in Lemma \ref{lemma:psd} by proving the existence of the inverse matrix of $\frac{1+\gamma}{\gamma} (\frac{\lambda}{\gamma} \bfL + \Gamma) - \bfR^T \bfR $. 

\begin{lemma} \label{lemma:psd}
	The inverse of matrix $\frac{1+\gamma}{\gamma} (\frac{\lambda}{\gamma} \bfL + \Gamma) - \bfR^T \bfR $ exists if $\sum_{i=1}^N r_{i,k} > 0, \forall k$, where $\Gamma = \emph{\textrm{diag}}( \bfone^T \bfR )$ and Laplacian matrix over a tree encoded in $\bfS$ is $\bfL = \emph{\textrm{diag}}(\bfS \bfone) - \bfS$.
\end{lemma}

%

The conditions $\sum_{i=1}^N r_{i,k} > 0, \forall k$, always hold in the case of the soft-assignment obtained by Proposition \ref{prop:mean-shift}.

\begin{proposition} \label{prop:ddr-WZY}
	By fixing $\{\bfS, \bfR\}$, problem (\ref{op:dimred-kmeans-relax-tree}) with respect to $\{ \bfW, \bfZ, \bfC \}$ has the following analytical solution:
	
	\begin{small}\vspace{-0.1in}
		\begin{align}
		\bfW = \bfU(:,1:d), ~~
		\bfZ = \bfQ \bfY \bfW, ~~
		\bfC = \left(\frac{\lambda}{\gamma} \bfL + \Gamma \right)^{-1} \bfR^T \bfZ
		\end{align}
	\end{small}\noindent
	where $\bfQ=\frac{1}{1+\gamma} \left[ \bfI_N + \bfR \left( \frac{1+\gamma}{\gamma} \left(\frac{\lambda}{\gamma} \bfL + \Gamma \right) - \bfR^T \bfR \right)^{-1} \bfR^T  \right] $, $\bfU$ and $\emph{\textrm{diag}}(\Lambda)$ are the eigenvectors and eigenvalues of matrix $\bfY^T \bfQ \bfY$ with $\emph{\textrm{diag}}(\Lambda)$ sorted in a descending order, respectively, $\Gamma = \emph{\textrm{diag}}( \bfone^T \bfR )$ and the Laplacian matrix over a tree encoded in $\bfS$ is defined as $\bfL = \emph{\textrm{diag}}(\bfS \bfone) - \bfS$.
\end{proposition}

By fixing $\{ \bfW, \bfZ, \bfY \}$, problem (\ref{op:dimred-kmeans-relax-tree}) with respect to $\{\bfS, \bfR\}$  is jointly convex optimization problem with respect to $\bfS$ and $\bfR$. Importantly, the subproblems with respective to $\bfS$ and $\bfR$ can be solved independently. According to Lemma \ref{prop:R}, the optimum $\bfR$ is given by equation (\ref{op:2-R}).
To obtain the optimum $\bfS$, the optimization problem with respect to $\bfS$, i.e., $ \min_{\bfS \in \S_{\T}} \sum_{ k,k' } s_{i,j}  ||\bfc_{k} -\bfc_{k'}||^2$, can be solved by Kruskal's method.

As discussed in Section \ref{sec:drtree-begin}, PCA is a special case of DRTree, so variable $\bfZ$ can be naturally initialized by PCA. 
By Proposition \ref{prop:mean-shift}, we can set $K=N$ and initialize $\bfY = \bfZ$. The pseudo-code of Discriminative DRTree is given in Algorithm \ref{algo2}, briefly named as \textit{DDRTree}. The implementations of {\it DRTree} and {\it DDRTree} in both MATLAB and R can be freely available\footnote{http://liwang8.people.uic.edu/}.

\section{Connections to Existing Methods} \label{sec:connect}

We have developed several methods based on regularized empirical Bayesian inference so that both the global assumption of generative model and the local assumption of manifold learning are naturally incorporated into a unified model. In addition to the relationships of our method to MVU and various probabilistic models such as PPCA and GPLVM discussed in the related work, we further present a detailed discussion of other existing methods that are closely related to our proposed model.

\subsection{Connection to Reversed Graph Embedding}

Reversed graph embedding \cite{Mao2015} was proposed to learn a set of principal points in the original space. Given a dataset $\D = \{\bfx_i\}_{i=1}^N$, it formulates the following optimization problem to learn a set of latent variables $\{ \bfz_1,\ldots, \bfz_N \}$ in low-dimensional space with $\bfz_i \in \Z$ given by

\begin{small}\vspace{-0.15in}
	\begin{align}
	\min_{\G \in \widehat{\G}_b } \min_{f_{\G} \in \mathcal{F}} \min_{\{\bfz_1,\ldots,\bfz_N\}} &~ \sum_{i=1}^N ||\bfx_i - f_{\G}(\bfz_i)||^2  \label{op:dimred-graph}\\
	&+ \frac{\lambda}{2}  \sum_{ (V_i,V_j) \in \E } b_{i,j} || f_{\G}(\bfz_i) - f_{\G}(\bfz_j) ||^2, \nonumber 
	\end{align}
\end{small}\noindent
where $\lambda \geq 0$ is a  parameter that controls the trade-off between the data reconstruction error and the objective function of reverse graph embedding, and $\widehat{\G}_b$ is a feasible set of graphs with the set $\V$ of vertices and a set $\E$ of edges specified by a set $\{b_{i,j}\}$ of edge weights. We consider learning a function $f_{\G} \in \mathcal{F}$ and $f_{\G}: \Z \rightarrow \mathcal{X}$ over $\G=(\mathcal{V}, \mathcal{E})$ that maps the intrinsic space $\Z$ to the input space $\mathcal{X}$. 

For simplicity, the work \cite{Mao2015} consider learning $f_{\G}(\bfz_i)$ as one single variable for variables $f_{\G}$ and $\bfz_i$.
In contrast, we in this paper aim to learn a set of points and the projection matrix as two separate variables, so that we can control the reduced dimensionality of the intrinsic space where the graph structure may reside. Moreover, we provide a general similarity matrix learning framework (\ref{op:ebi-SW}) for principal graph learning and special tree structure learning formulations (\ref{op:spl-graph}) and (\ref{op:dimred-kmeans-tree}).

\subsection{Connection to Maximum Entropy Unfolding}

MEU \cite{lawrence2012unifying} was proposed to directly model the density of observed data $\mathbf{Y} = [\mathbf{y}_1,\ldots, \mathbf{y}_N]$ by minimizing the KL divergence between a base density  $m(\mathbf{Y})$ and the density $p(\mathbf{Y} )$ given by
$\min_{p(\mathbf{Y})} \int p(\mathbf{Y} ) \log ( {p(\mathbf{Y} )}/{ m(\mathbf{Y}) } )$,
under the constraints on the expected squared inter-point distances $\phi_{i,j}$ of any two samples, $\mathbf{y}_i$ and $\mathbf{y}_j$. Let $m(\mathbf{Y})$ be a very broad, spherical Gaussian density with covariance $\lambda^{-1} \mathbf{I}$. The density function is then constructed as 

\begin{small}\vspace{-0.1in}
	\begin{align*}
	p(\mathbf{Y}) \propto \exp \Big( -\frac{1}{2} \textrm{Tr}(\lambda \mathbf{Y} \mathbf{Y}^T) \Big) \exp \Big( -\frac{1}{2} \sum_{i} \sum_{j \in \mathcal{N}_i} w_{i,j} \phi_{i,j}   \Big),
	\end{align*}
\end{small}\noindent
even though the explicit form of these constraints is not given. 
The Laplacian matrix $\mathbf{L}$ defined over similarity $w_{i,j}$ is achieved by maximizing the logarithmic function of $p(\mathbf{Y})$. Finally, the embedding is obtained by applying KPCA on the kernel matrix $\mathbf{K} = (\mathbf{L} + \lambda \mathbf{I}_N )^{-1}$. 

One of the key differences is that our framework directly models the posterior distribution $p(\mathbf{X} | \mathbf{Y} )$ of latent data, while MEU models the density of observed data. As a result, MEU has to assume that the data features are i.i.d. given the model parameters. However, this assumption is hardly satisfied if feature correlation exists. In contrast, our model assumes that the reduced features in the latent space are i.i.d, which is more reasonable than that used in MEU since the latent space is generally assumed to be formed by a set of orthogonal bases, such as PCA and KPCA.

\subsection{Connection to Structure Learning}


SMCE \cite{elhamifar2011sparse} was proposed using $\ell_2$ norm over the errors that measure the linear representation of every data point by using its neighborhood information. 
Similarly, $\ell_1$ graph was learned for image analysis using $\ell_1$ norm over the errors for enhancing the robustness of the learned graph \cite{Cheng2010}. 
These two methods  \cite{Cheng2010,elhamifar2011sparse} learn a directed graph from data so that they might yield suboptimal results by heuristically transforming a directed graph to an undirected graph for clustering and dimensionality reduction.

Instead of learning directed graphs by using the above two methods, an integrated model for learning an undirected graph by imposing a sparsity penalty (i.e., $\ell_1$ prior) on a symmetric similarity matrix and a positive semidefinite constraint on the Laplacian matrix was proposed \cite{lake2010discovering}, given by,

\begin{small}\vspace{-0.15in}
	\begin{align}
	\max_{\bfQ, \bfS, \sigma^2}&~ \log\det(\bfQ) - \frac{1}{D}\Tr( \bfQ \bfY \bfY^T ) - \frac{\beta}{D} ||\bfS||_1 \label{op:sl}\\
	\textrm{s.t.} &~ \bfQ = \textrm{diag}(\bfS \bfone) - \bfS + \bfI_N / \sigma^2 \nonumber\\
	&~s_{i,i} = 0, s_{ij}\geq 0, i=1,\ldots, N \nonumber\\
	&~\sigma^2 > 0 \nonumber
	\end{align}	
\end{small}\noindent
where $\beta>0$ is a regularization parameter.
Another approach dimensionality reduction through regularization of the inverse covariance in the loglikelihood (DRILL) \cite{lawrence2012unifying} was proposed by applying an $\ell_1$ prior to the elements of an inverse covariance, given by,

\begin{small}\vspace{-0.15in}
	\begin{align}
	\max_{\Lambda}& \frac{D}{2}\log\det(\Lambda + \lambda \bfI_N) - \frac{1}{2}\Tr( (\Lambda + \lambda \bfI) \bfY \bfY^T ) - ||\Lambda||_1, \label{op:drill}
	\end{align}	
\end{small}\noindent
and an implied covariance matrix is $\bfK = (\Lambda + \lambda \bfI_N )^{-1}$.

By comparing (\ref{op:ebi-SW}) with (\ref{op:sl}) and (\ref{op:drill}), we can see that (\ref{op:ebi-SW}) is more similar than (\ref{op:sl}) instead of (\ref{op:drill}) since the sparsity is imposed on $\bfS$, not on $\Lambda$, which is analogous to the Laplacian matrix $\bfL = \textrm{diag}(\bfS \bfone) - \bfS $. In fact, the sparisties of $\bfS$ and $\bfL$ are the same except the diagonal, but the properties of two matrices are very different. Our model demonstrates two key differences from the two methods. First, (\ref{op:ebi-SW}) has additional term for modeling the data generation process. If $\gamma=0$ and the absolute difference is used (see Section \ref{sec:expect-distance} ), the primal problem of (\ref{op:ebi-SW}) is equivalent to ($\ref{op:sl}$). $\gamma>0$ is useful to retain the structure of data after dimensionality reduction. Second, our structure learning model DRTree and DDRTree takes the spanning trees as the candidate structure, which is very different from $\ell_1$ regularization over $\bfS$ so that our structure learning methods can transform original points into embedded points that form spanning trees in the latent space. 


%
%

\section{Experiments} \label{sec:experiments}

We perform extensive experiments to verify our proposed models, SPL and DDRTree, separately, by comparing them to various existing dimensionality reduction methods on a variety of synthetic and real world datasets.

\subsection{Structured Projection Learning} \label{sec:SPL}

\begin{figure}[t]
	\centering
	\includegraphics[width=1.1\textwidth]{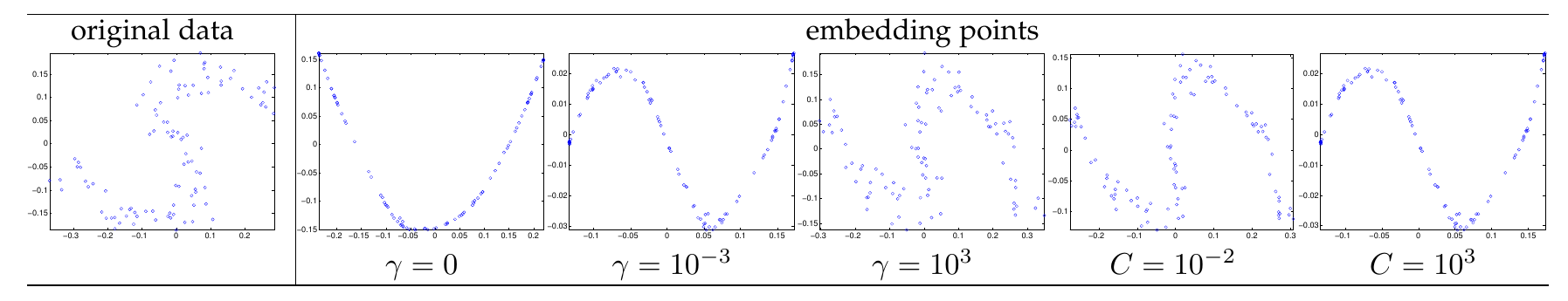}
	\caption{The visualization results for parameter sensitivity analysis of SPL on DistortedSShape by either varying $\gamma \in \{0, 10^{-3}, 1 \}$ with prefixed $C=\infty$, or varying $C \in \{10^{-2}, 10^3, \infty\}$ with $\gamma=10^{-3}$ using the neighborhood size as $10$ in $2$-D space. } \label{fig:sensitivity}
	\vspace{-0.2in}
\end{figure}

\subsubsection{Parameter Sensitivity Analysis}

We investigate the parameter sensitivity of the proposed SPL method by varying $\gamma$ and $C$ on DistortedSShape, a synthetic data  of $100$ data points, which has been used in \cite{Kegl2000}. For simplicity, we study the influence of parameters by varying one and fixing the other. The neighborhood size is set to $10$ and the reduced dimension is $2$. We vary $\gamma \in  \{0, 10^{-3}, 1 \}$ and $C \in \{10^{-2}, 10^3, \infty\}$. 

Fig. \ref{fig:sensitivity} shows the original data and  the resulting embeddings using SPL by varying $\gamma$ and $C$. We made the following observations from Fig. \ref{fig:sensitivity}: (i) SPL with a small $\gamma$ can obtain embedding points with less noises (smooth S-shape skeleton); (ii) SPL with $\gamma=0$ changes the spatial intrinsic structure from S-shape to V-shape structure, while the true structures can be retained by introducing the generative model using a small $\gamma$; (iii) a large $C$ is preferred since it does not allow a large violation of pairwise distances from original data to the embedding data; (iv) SPL can achieve relatively good embeddings in a large interval of $\gamma$ and $C$. All these observations are consistent with algorithmic analysis in Section \ref{sec:aa}. 

\begin{figure*}[!htbp]
	\centering
	\includegraphics[width=1.1\textwidth]{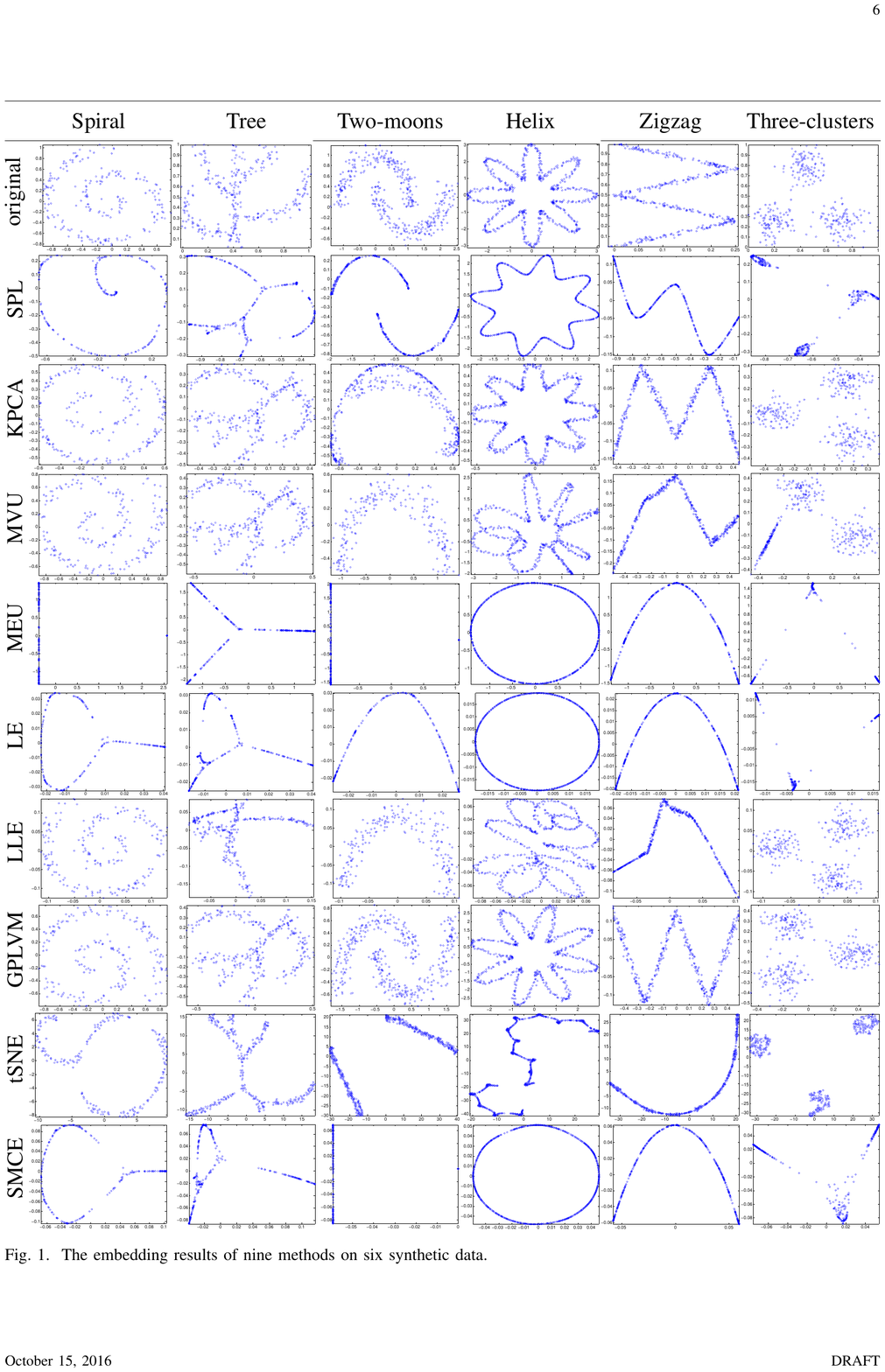}
	\caption{The visualization results of embedding points using six methods on six synthetic data. Due to the space limit, the visualization results of the remaining methods are reported in the supplementary materials.}\label{tab:synthetic}
\end{figure*}

\subsubsection{Synthetic Data} \label{sec:synthetic}

We conduct experiments for data visualization on six synthetic datasets by comparing the proposed SPL method with various dimensionality reduction methods, including KPCA \cite{Scholkopf1999}, MVU \cite{weinberger2004learning}, MEU \cite{lawrence2012unifying}, LE \cite{belkin2001laplacian}, LLE \cite{saul2003think}, GPLVM \cite{lawrence2005probabilistic}, tSNE \cite{van2008visualizing}, SMCE \cite{elhamifar2011sparse}.
Two datasets, \textit{Spiral} and \textit{Zigzag}\footnote{https://www.lri.fr/$\sim$kegl/researchUdeM/research/pcurves\\/implementations/Samples/}, are used, which are also used in the principal curve learning \cite{Kegl2000} and the underlying structures are spiral and zigzag curves, respectively.
Dataset \textit{2moons}\footnote{manifold.cs.uchicago.edu/manifold\_regularization/2moons.mat} is used in manifold-based semi-supervised learning \cite{belkin2006manifold} for distinguishing two classes, so the true skeleton structure consists of two smooth curves.
Following the work \cite{van2009dimensionality}, we generate a 3-D datasets, \textit{helix}  by using drtoolbox, where the data point $\mathbf{y}_i$ is computed by
$\mathbf{y}_i = [ (2 + \cos(8 p_i))\cos(p_i), (2 + \cos(8 p_i))\sin(p_i), \sin(8 p_i)]$ where $p_i$ is a random number that is sampled from a uniform distribution with support $[0,1]$. Two datasets, \textit{Tree} and \textit{Three-clusters} are from \cite{yao2015feature}, where the underlying structures are tree and three clusters with weak connections, respectively. Each data is projected to a $2$-D space for data visualization by the above mentioned methods. Gaussian kernel is used for kernel-based methods or methods with prefixed similarity measurement. We set $C=10^3$ and $\gamma=10^{-3}$ for SPL for all synthetic datasets. The same neighborhood size parameter $K=10$ is used to construct neighborhood graph for neighborhood based methods. 

Fig. \ref{tab:synthetic} shows the visualization results of embedding points in 2-D space obtained by nine methods on six synthetic datasets. 
We made the following observations. Some methods, such as KPCA, MVU, LLE and GPLVM, cannot tolerate noise of data so that the learned embeddings do not show a smooth skeleton structure. Some methods like MEU, LE, tSNE and SMCE can learn a smooth skeleton structure from noise data, but most of them are not consistent with the underlying structures. 
For example, only tSNE and SPL can obtain two separate curves from two-moons data, but the curves learned by SPL are much smoother and more similar to moon-shaped structures than tSNE. On helix and zigzag, most methods obtain the embeddings that are very different from the true skeleton structures.
Only SPL can obtain smooth skeleton structures by maintaining the underlying structures on all six datasets. These observations imply that SPL is able to learn a smooth skeleton structure from noisy data, and simultaneously retain the underlying structure of data in 2-D space.

\subsubsection{Visualization of Two Real Datasets} \label{sec:embdding-teapot}

\begin{figure}
	\centering
	\includegraphics[width=1.0\textwidth]{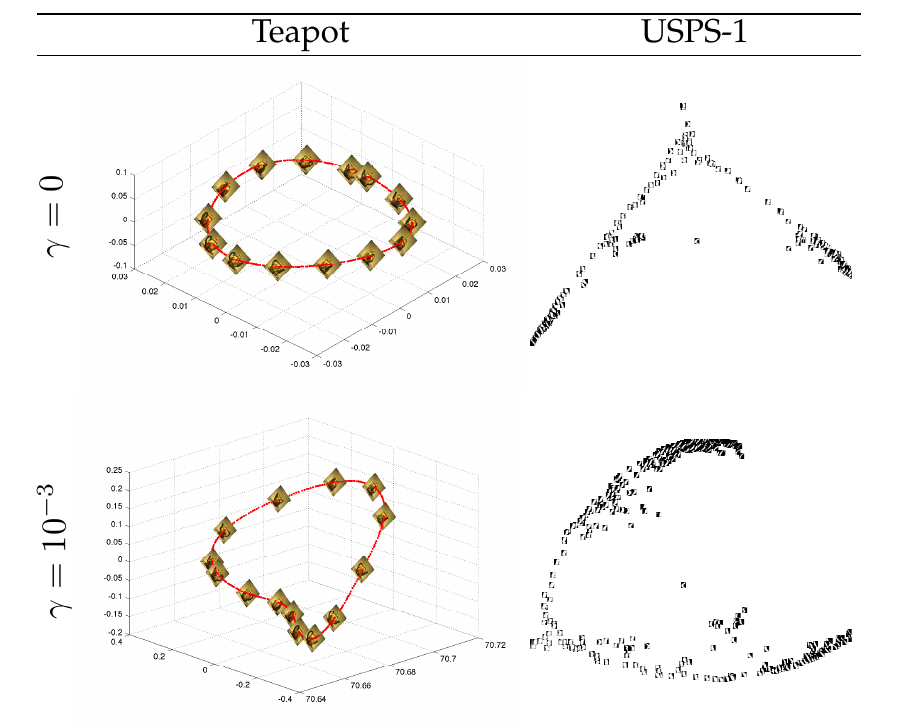}
	\caption{Visualization results of our proposed method with parameters $\gamma\in \{ 0, 10^{-3} \}$ and $C=10^3$ on Teapot and USPS with tag 1.}\label{fig:real-visual} 
\end{figure}

A collection of $400$ teapot images from \cite{Weinberger2006} are used\footnote{http://www.cc.gatech.edu/$\sim$lsong/data/teapotdata.zip} for the purpose of visualization.  These images are taken successively as a teapot was rotated $360^{\circ}$. Our goal is to unveil the circular structure in $3$-D space that organizes the 400 images.  Each image consists of $76 \times 101$ RGB pixels and is represented as a vector of size $23,028$. Thus, this data is high-dimensional. We run our algorithm with parameters $\gamma\in \{ 0, 10^{-3} \}$ and $C=10^3$. The visualization results are reported in Fig. \ref{fig:real-visual}. A two-dimensional representation of the same set of teapot images is given in \cite{Weinberger2006}, where MVU can also successfully arrange these images in a circle (see Figure 3 in \cite{Weinberger2006}). This is the same as our result with $\gamma = 0$. However, when $\gamma = 10^{-3}$, the visualization of embeddings is slightly different, although the similar circular structure can be achieved.

Another data is USPS handwritten digits\footnote{http://www.cs.nyu.edu/$\sim$roweis/data/usps\_all.mat}, which contains handwritten digits from 0 to 9 with different written styles.  Each one is a gray image of size $16\times 16$. The vectorization of each image with label ``$1$'' is used to study the embeddings in $2$-D space since these images demonstrate very clear written styles. There are $1,100$ images in total. Following the same setting as did for teapot data, the visualization results of USPS data are shown in Fig. \ref{fig:real-visual}, where each image is shown in the position located by its associated embedded point in 2-D space. We observe that the images of ``$1$"s are sorted in an order such that the angular degree of image ``$1$''s is changed continuously, and the perpendicular ones are shown in the middle of the learned skeleton structure. The visualization difference between $\gamma=0$ and $\gamma=10^{-3}$ is that the latter demonstrates a smoother change than the former since the structure of the latter is an arc while the structure of the former is a right angle. 

\subsubsection{Classification Performance of Embeddings}


\begin{landscape}
	\begin{table*}[!htbp]
		\centering
		\caption{The leave-one-out cross validation accuracy of one-nearest neighbor classifier over ten datasets. $N$ is the number of data points. $c$ is the true number of clusters. $D$ is the original dimensionality and $d$ is the reduced dimensionality. The best results are in bold.} \label{tab:clustering}
		\vspace{-0.1in}
		\begin{small}
			\begin{tabular}{@{}l@{}|cccccccccc@{}} \hline
				& Iris & CMU-PIE & COIL20 & Isolet & Pendigits & Satimage & USPS & Vehicle & Segment & Letter \\\hline
				$(N, c)$ & (150, 3) & (3329, 68) & (1440, 20) & (3119, 2) &(3498, 10) &(4435, 6) & (2007, 10) &(846, 4) &(231,7) &(5000, 26)\\
				$(D, d)$ & (4, 2) &(1024, 39) &(2014,84) &(617,165) &(16, 9) &(36, 6) &(256, 32) &(18, 6) &(19,7)&(16,12)\\\hline
				LLE \cite{saul2003think}&0.9467	&0.9655	&0.9965	&0.9298	&0.9760	&0.8570	&0.9013	&0.6537	&0.9623	&0.8960 \\
				LE \cite{belkin2001laplacian} &0.9133	&0.6248	&0.9833	&0.9368	&0.9714	&0.8586	&0.9023	&0.5981	&0.9398	&0.7436 \\
				MVU \cite{weinberger2004learning}&0.6533	&0.4662	&0.7660	&0.8035	&0.9737	&0.8607	&0.7693	&0.5579	&0.9342	&0.6042 \\
				KPCA\cite{Scholkopf1999} &0.9000	&0.2701	&0.5583	&0.7086	&0.9883	&0.8462	&0.3303	&0.5615	&0.9550	&0.8488 \\
				GPLVM \cite{lawrence2005probabilistic}&0.9333	&0.9787	&\textbf{1.0000}	&0.8410	&0.9866	&0.8884	&0.5944	&0.5898	&0.9688	&0.8974 \\
				MEU \cite{lawrence2012unifying}&0.8867	&0.9507	&\textbf{1.0000}	&0.9349	&0.9840	&0.8652	&0.9312	&0.6407	&0.9537	&0.1244 \\
				SMCE \cite{elhamifar2011sparse}&0.9400	&0.9612	&\textbf{1.0000}	&0.9314	&0.9806	&0.8848	&0.9307	&0.6832	&0.9398	&0.8790	\\
				tSNE \cite{van2008visualizing}&\textbf{0.9600}	&0.9751	&\textbf{1.0000}	&0.9468	&0.9909	&\textbf{0.9037}	&0.9292	&0.6738	&0.9671	&\textbf{0.9134} \\
				SPL &\textbf{0.9600}	&\textbf{0.9805}	&\textbf{1.0000}	&\textbf{0.9497}&\textbf{0.9943}&0.8938	&\textbf{0.9317}	&\textbf{0.6915}&\textbf{0.9680}	&0.9054 \\\hline
			\end{tabular} \vspace{-0.1in}
		\end{small}
	\end{table*}
\end{landscape}

As shown in Table \ref{tab:clustering}, ten datasets taken from the UCI and Statlib repositories are used to evaluate classification performance of embedded points learned by baseline methods same as those used in the experiments on synthetic data. The reduced dimensionality of data is shown in Table \ref{tab:clustering} by preserving $95\%$ of energy of data. Following \cite{weinberger2004learning}, we use the leave-one-out cross validation accuracy as the criterion for evaluating one-nearest neighbor classifier on the embeddings learned by these baseline methods. For methods that require $K$ nearest neighbor graph as the input, we tune $K \in [5, 10, 15, 20, 30, 50]$. We tune the parameter $\lambda \in [0.01, 0.1, 1, 10]$ for SMCE. Other parameters are set as the default values in the drtoolbox \footnote{https://lvdmaaten.github.io/drtoolbox/}. In addition, we tune $C=[10, 10^3]$ and $\gamma\in [0, 10^{-3}]$. The best results are reported for every baseline methods by tuning their own parameters.

Table \ref{tab:clustering} shows the leave-one-out cross validation accuracy of one-nearest neighbor classifier over the embeddings learned by nine methods on ten benchmark datasets. It is clear to see that SPL is competitive to tSNE in terms of classification accuracy, and demonstrates much better than the others. As shown in \cite{van2008visualizing}, tSNE helps to achieve good classification performance by learning a new embedding of original data. The learning criterion of tSNE is better suitable for clustering/classification, but it is not appropriate for learning skeleton structures in a latent space as observed in Section \ref{sec:synthetic} and Section \ref{sec:embdding-teapot}. These results imply that SPL is not only suitable for learning skeleton structures in latent spaces from high-dimensional data, but also it can achieve competitive or better classification performance on the learned embedding points by comparing with various existing methods.

\subsection{Learning Tree Structures From Real Datasets}

We investigate the ability of our proposed DDRTree method to automatically discover tree structures from three real-world datasets. The latent tree structures of these three datasets include principal curves, hierarchical tree structures, and a cancer progression path.

\subsubsection{Principal Curve}

The same teapot data in Section \ref{sec:embdding-teapot} is used to justify our tree structure learning methods.
Similar to \cite{Song2007}, the data in each dimension is normalized to have zero mean and unit standard deviation. And, a kernel matrix ${\bf Y}$ is generated where $Y(i,j) =  \exp(- ||\bfy_i - \bfy_j||^2 /D)$.

\begin{figure}[ht]
	\centering
	\begin{tabular}{cc} 
		\includegraphics[width=0.5\textwidth]{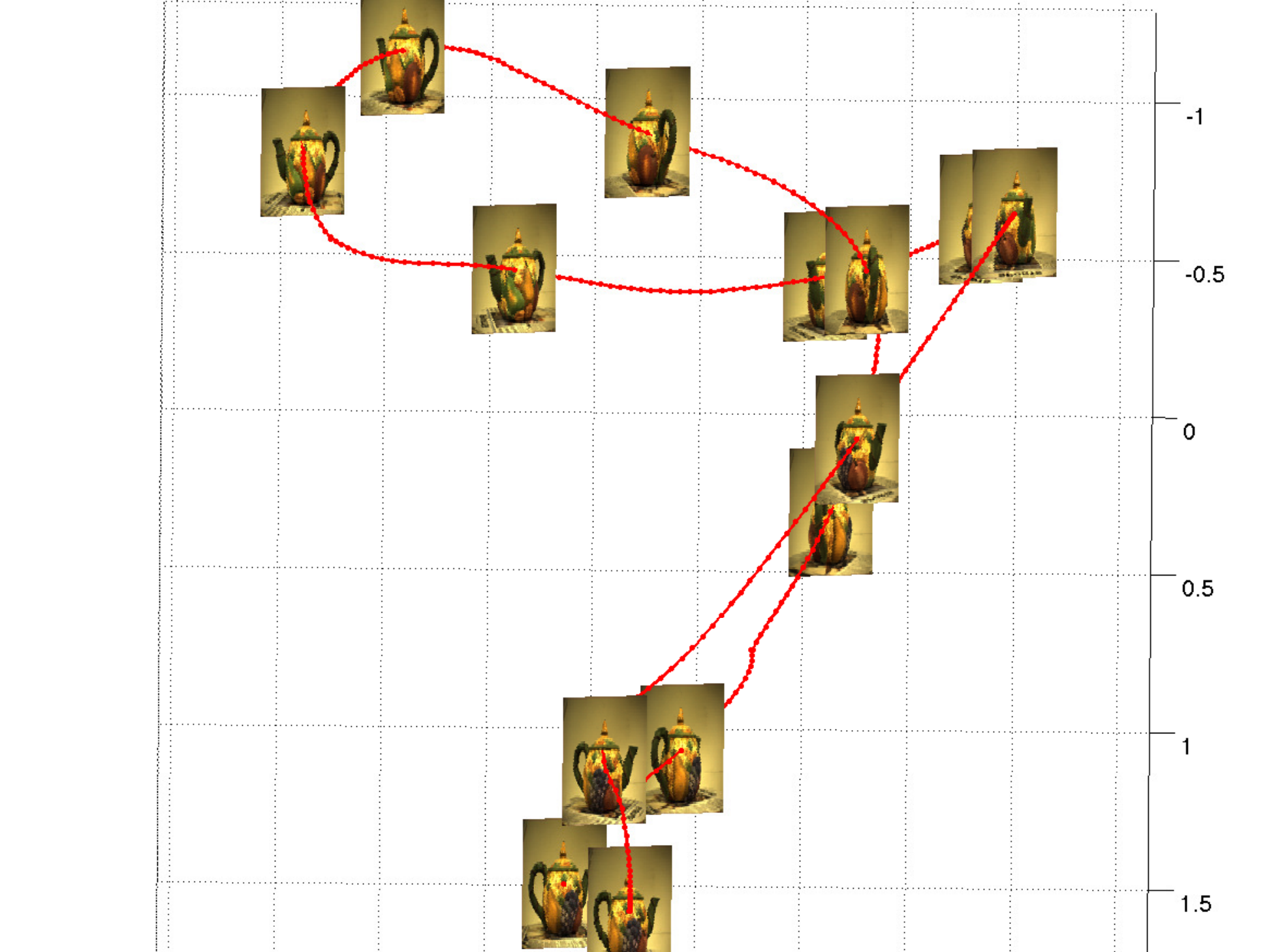} & 
		\includegraphics[width=0.5\textwidth]{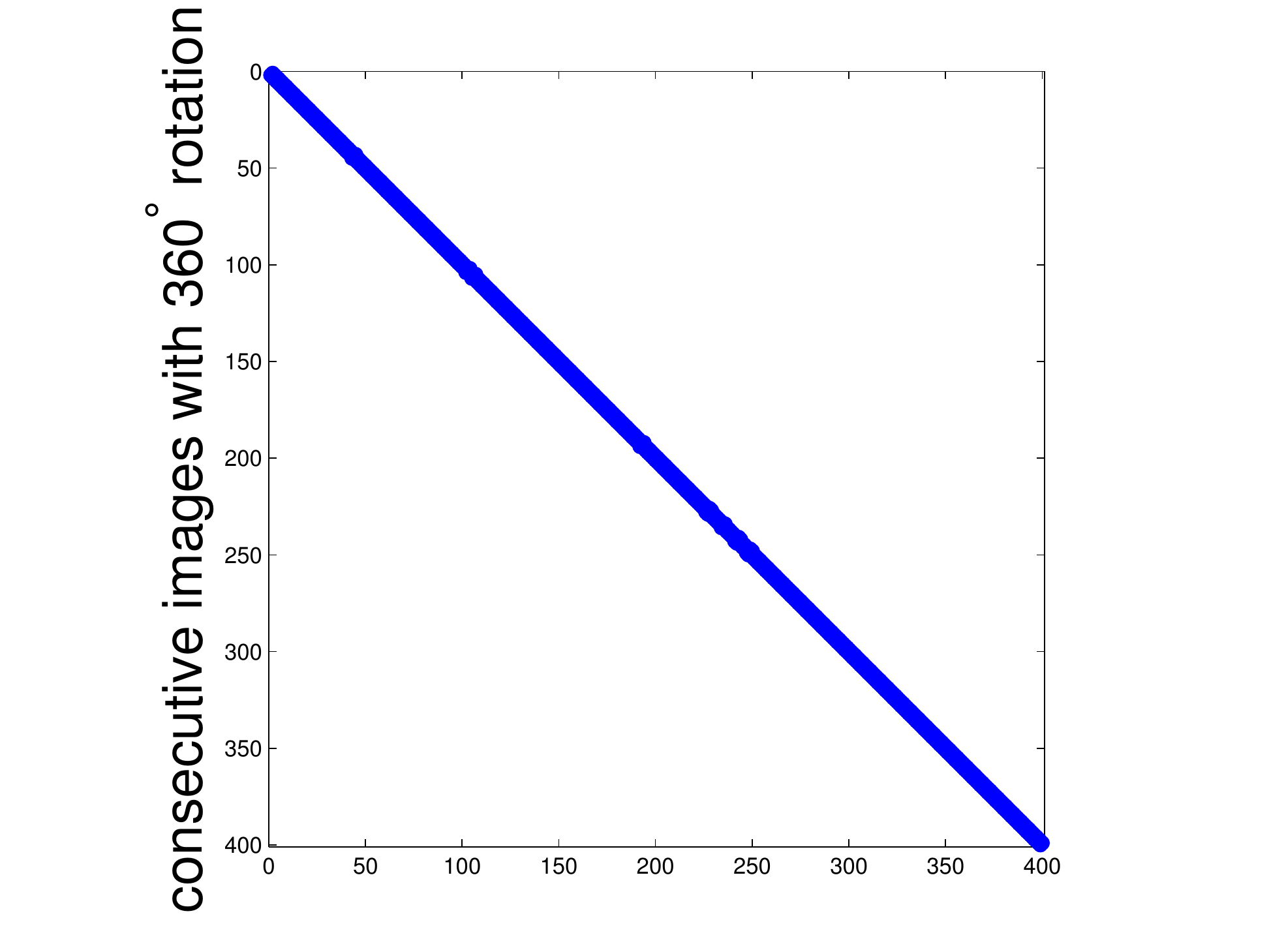}\\
		(a) curve &
		(b) adjacency matrix
	\end{tabular}
	\caption{Experimental results of DDRTree applied to Teapot images. (a) principal curve generated by DRTree. Each dot represents one teapot image. Images following the principal curve are plotted at intervals of $30$ for visualization. (b) The adjacency matrix of the curve follows the ordering of the $400$ consecutive teapot images with $360^\circ$ rotation. } \label{fig:teapot}
	\vskip -6pt
\end{figure}

We run our proposed DDRTree method using the kernel matrix as the input. We set $\lambda=0.1 \times N$ and $d=36$ that keeps $95\%$ of total energy. The experimental results of DDRTree are shown in Figure \ref{fig:teapot}. The principal curve (Figure \ref{fig:teapot}(a)) is shown in terms of the first $3$ columns of the learned projection matrix $\bfW$ as the coordinates where each dot represents one image. The sampled images at intervals of $30$ are plotted for the purpose of visualization. Figure \ref{fig:teapot}(b) shows the linear chain dependency among teapot images following the consecutive rotation process. We can see that the curve generated by our method is consistent with the rotating process of the $400$ consecutive teapot images. 

A similar result is also recovered by CLUHSIC, which assumes that the label kernel matrix is a ring structure \cite{Song2007}. However, there are three main differences. First, we learn a projection space where images are arranged in the form of a principal curve, while CLUHSIC applies KPCA to transform original data to an orthogonal space where clustering is performed. Second, the principal curve generated by our DDRTree method is much smoother than that obtained by CLUHSIC (see Figure 4 in \cite{Song2007}). Third, our method learns the adjacency matrix from the given dataset, but CLUHSIC requires a label matrix as {\em a prior}.  We attempted to run MVU by keeping $95\%$ energy, i.e., $d=36$. However, storage allocation fails due to the large memory requirement of solving a semidefinite programming problem in MVU. Hence, MVU fails to learn a relatively large intrinsic dimensionality. However, our method does not have this issue.

\subsubsection{Hierarchical Tree}

Facial expression data\footnote{http://www.cc.gatech.edu/$\sim$lsong/data/facedata.zip} is used for hierarchical clustering, which takes into account both the identities of individuals and the emotion being expressed \cite{Song2007}. This data contains $185$ face images ($308 \times 217$ RGB pixels) with three types of facial expressions (NE: neutral, HA: happy, SO: shock) taken from three subjects (CH, AR, LE) in an alternating order, with around $20$ repetitions each. Eyes of these facial images have  been aligned, and the average pixel intensities have been adjusted. As with the teapot data, each image is represented as a vector, and is normalized in each dimension to have zero mean and unit standard deviation. 


\begin{figure}[t]
	\centering
	\begin{tabular}{cc}
		\includegraphics[width=0.5\textwidth]{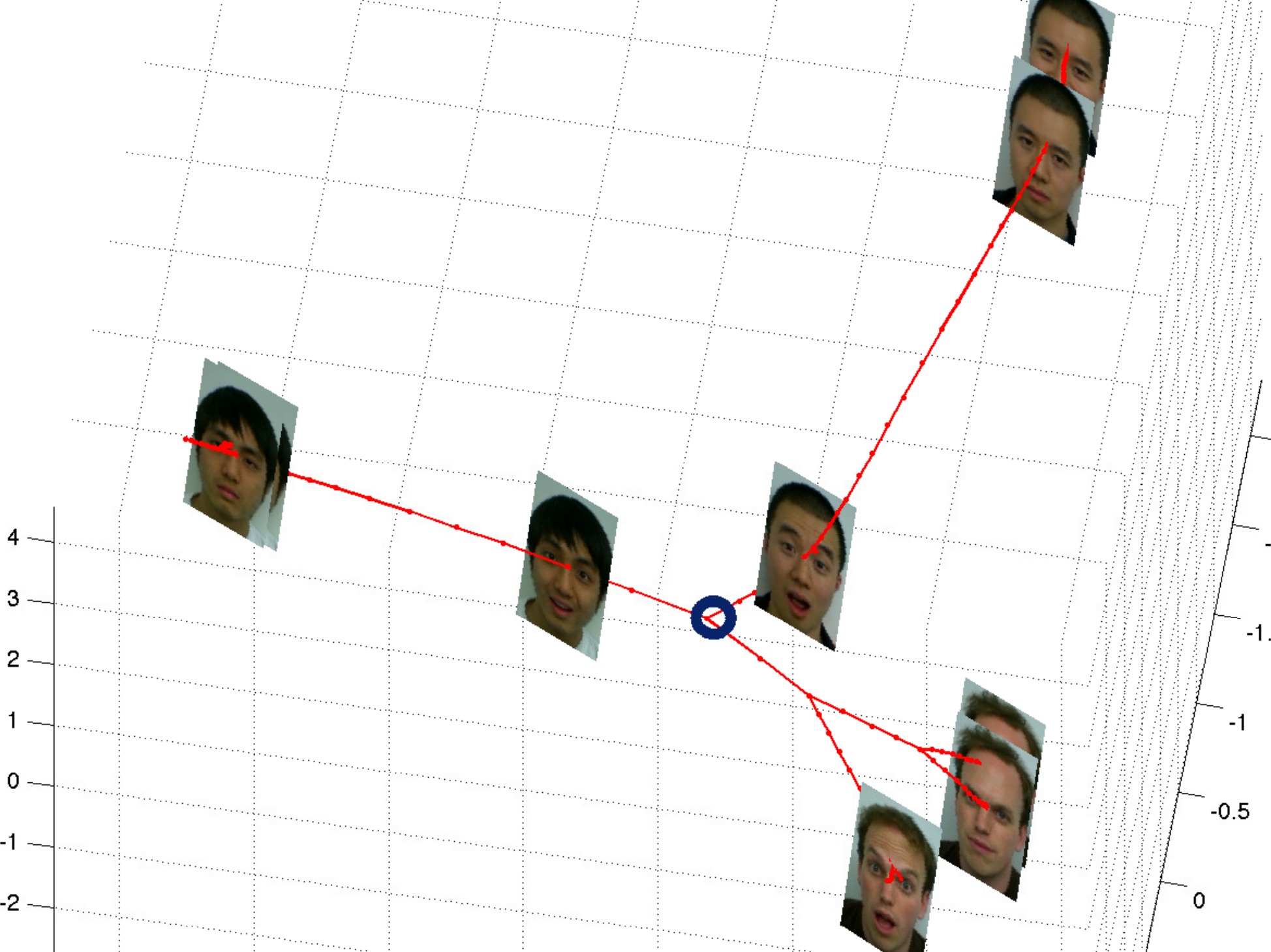} 
		& \includegraphics[width=0.5\textwidth]{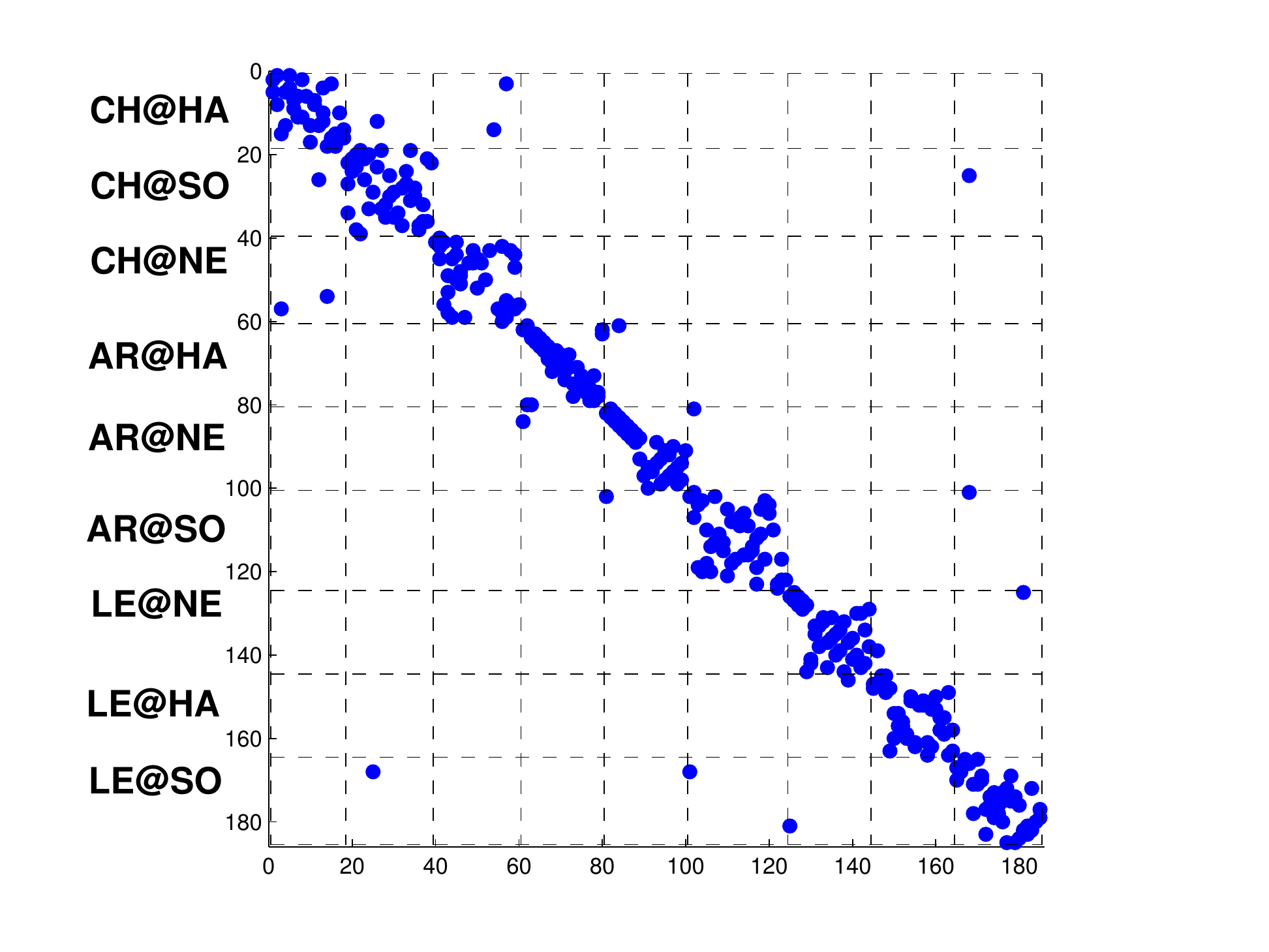} \\
		\!\!\!\!\!\!(a) tree & (b) adjacency matrix
	\end{tabular}
	\caption{Experimental results of our DDRTree method performed on facial expression images. (a) A hierarchical tree generated by DRTree. Each dot represents one face image. Images of three types of facial expressions from three subjects are plotted for visualization. The black circle is the root of the hierarchical structure; (b)  The adjacency matrix of the tree on nine blocks indicates that each block corresponds to one facial expression of one subject. } \label{fig:face}
	\vskip -6pt
\end{figure}

A kernel matrix is used as the input to DDRTree. $\lambda=0.1 \times N$ and $d=185$. The experimental results are shown in Figure \ref{fig:face}. We can clearly see  that three subjects are connected through different branches of a tree.  If we take the black circle in Figure \ref{fig:face}(a) as the root of a hierarchy, the tree forms a two-level hierarchical structure.  As shown in Figure \ref{fig:face}(b), all three facial expressions from three subjects are also clearly separated. A similar two-level hierarchy is also recovered by CLUHSIC (Figure 3(b) in \cite{Song2007}). However,  the advantages of using DDRTree discussed above for teapot images are also applied here.
In addition, we can observe more detailed information from the tree structure. For example, LE@SO is the junction to other two subjects, i.e., AR@SO and CH@SO, which can be observed from the $9$th row of the adjacency matrix \ref{fig:face}(b). This observation suggests that the shock is the most similar facial expression among three subjects. However, CLUHSIC is not able to obtain this information. 

\subsubsection{Cancer Progression Path} \label{sec:cancer}

\begin{figure}[t]
	\centering 
	\includegraphics[width=0.8\textwidth]{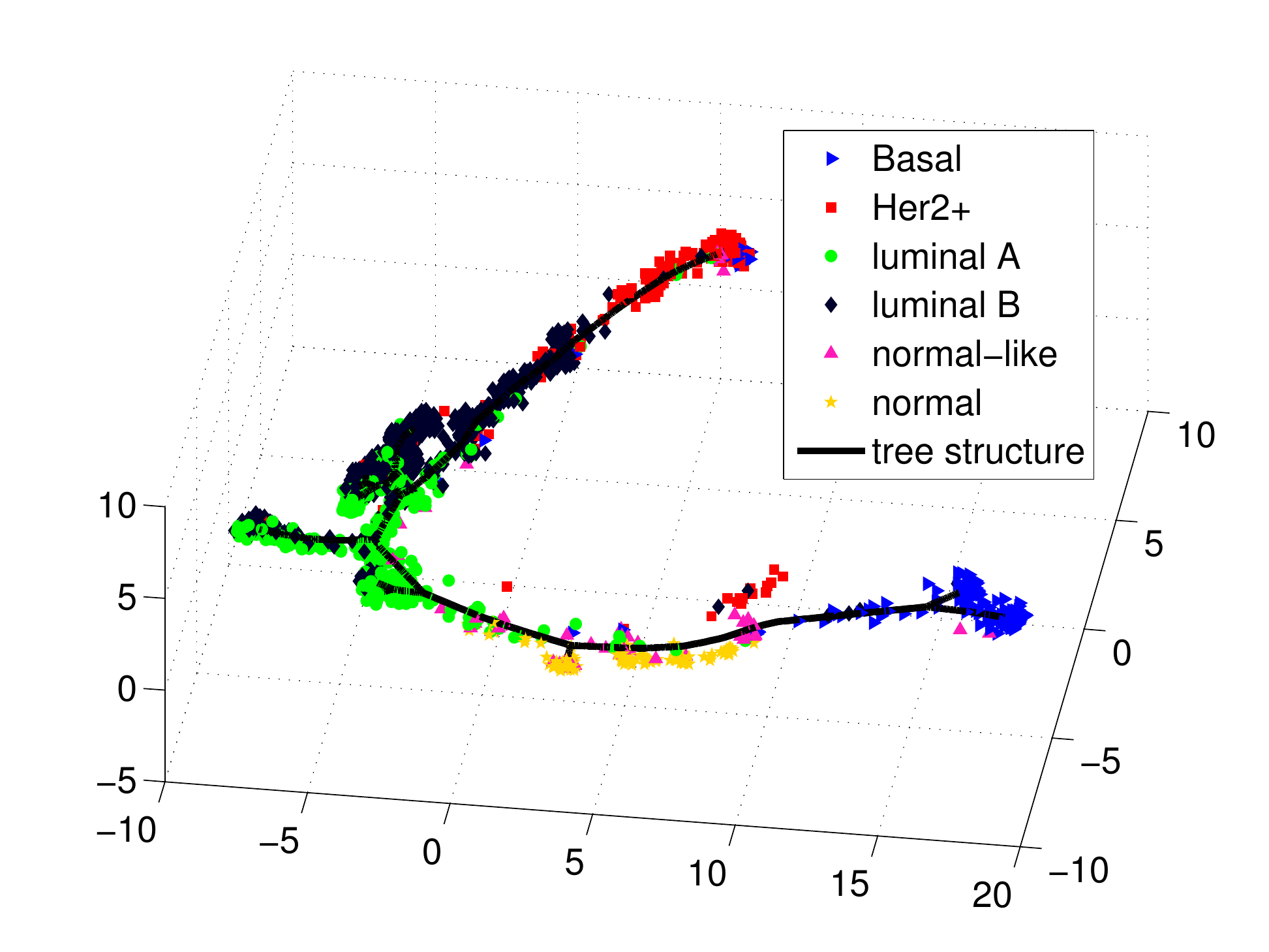} 
	\caption{Graph structure learned by DDRTree on breast cancer dataset with $d=80$ and visualized in three-dimensional space spanned by the first three components of the learned projection matrix.} \label{fig:cancer} 
\end{figure}

We are particularly interested in studying human cancer, a dynamic disease that develops over an extended time period. 
Once initiated from a normal cell, the advance to malignancy can to some extent be considered a Darwinian process - a multistep evolutionary process - that responds to selective pressure \cite{greaves2012clonal}. The disease progresses through a series of clonal expansions that result in tumor persistence and growth, and ultimately the ability to invade surrounding tissues and metastasize to distant organs. As shown in Figure \ref{fig:cancer}(a), the evolution trajectories inherent to cancer progression are complex and branching \cite{greaves2012clonal}. Due to the obvious necessity for timely treatment, it is not typically feasible to collect time series data to study human cancer progression [28]. However, as massive molecular profile data from excised tumor tissues (static samples) accumulates, it becomes possible to design integrative computation analyses that can approximate disease progression and provide insights into the molecular mechanisms of cancer.  We have previously shown that it is indeed possible to derive evolutionary trajectories from static molecular data, and  that breast cancer  progression can be represented by a high-dimensional manifold with multiple branches \cite{Sun2014}.

We interrogate a large-scale, publicly available breast cancer dataset \cite{Shah2012} for cancer progression modeling. The dataset contains the expression levels of over $25,000$ gene transcripts obtained from 144 normal breast tissue samples and $1,989$ tumor tissue samples.  By using a nonlinear regression method, a total of $359$ genes were identified that may  play a role in cancer development \cite{Sun2014}. In the analysis, we set $\lambda=5 \times N$ and $d=80$ that retains $90\%$ of energy.

Figure \ref{fig:cancer}(b) shows the learned latent structures and latent points in a reduced dimensional space. Each tumor sample is colored with its corresponding PAM50 subtype label, a molecular approximation that uses a 50-gene signature to group breast tumors into five subtypes including normal-like, luminal A, luminal B, HER2+ and basal \cite{Parker2009}. Basal and HER2+ subtypes are known to be the most aggressive breast tumor types. The learned graph structure in the low-dimensional space suggests a linear bifurcating progression path, starting from the normal tissue samples, and diverging to either luminal A or basal subtypes. The linear trajectory through luminal A continues to luminal B and to the HER2+ subtype. Significant side-branches are evident for both luminal A and luminal B subtypes, suggesting that these subtypes can be further delineated. The revealed data structure is consistent with the proposed branching architecture of cancer progression shown in Figure \ref{fig:cancer}.

\section{Conclusion} \label{sec:conclusion}

In this paper, we proposed a general probabilistic framework for dimensionality reduction, which not only takes the noise of data into account, but also utilizes the neighborhood graph as the locality information. Based on this framework,
we presented a model that can learn a smooth skeleton of embedding points from high-dimensional, noisy data. In order to learn an explicit graph structure, we developed another new dimensionality reduction method  that  learns a latent tree structure and low-dimensional feature representation simultaneously. We extended the proposed method for clustering problems by imposing the constraint that data points belonging to the same cluster are likely to be close along the learned tree structure. 
The experimental results demonstrated the effectiveness of the proposed methods for recovering intrinsic structures from real-world datasets. 
Dimensionality reduction via learning a graph is formulated from a general graph, so the development of new dimensionality reduction methods for various specific structure is also possible.

\appendix
\section{Proofs}
\subsection{Proof of Proposition 1}
In order to prove the results, we first derive the explicit expression of $q(\bfX)$ by taking the Lagrangian duality and obtaining the partial dual problem.

By fixing $\bfW$ and introducing dual variables $\bfS \in \bbR^{N \times N}$ with the $(i,j)$th element denoted as $s_{i,j} \geq 0$, the Lagrangian function $L(q(\bfX), \bfW, \bfS,\bm{\xi} )$ can be formulated as
\begin{small}
	\begin{align*}
	&\textrm{KL}(q(\bfX) || \pi(\bfX)) - \int \log p(\bfY | \bfX, \bfW, \gamma) q(\bfX) d \bfX + C ||\bm{\xi}||^2_F \\
	&+ \sum_{i,j \in \N_i} s_{i,j} \left(\bbE_{q(\bfX)} \Big[ ||\bfx_i - \bfx_j||^2 \Big] - \phi_{i,j} - \xi_{i,j} \right).
	\end{align*}
\end{small}\noindent
According to the duality theorem, we have the following KKT conditions,
\begin{small}
	\begin{align*}
	1 + \log ( q(\bfX) / \pi(\bfX) ) - \log p(\bfY | \bfX, \bfW, \gamma) + s_{i,j} ||\bfx_i - \bfx_j||^2 = 0 \\
	s_{i,j} \left(\bbE_{q(\bfX)} \Big[ ||\bfx_i - \bfx_j||^2 \Big] - \phi_{i,j} - \xi_{i,j} \right) = 0, \forall i, j\in\N_i \\
	2 C \xi_{i,j} - s_{i,j} = 0, \forall i, j\in\N_i.
	\end{align*}
\end{small}\noindent
The first equation leads to the following analytic form of posterior distribution
\begin{small}
	\begin{align*}
	q(\bfX) \propto \pi(\bfX) p(\bfY | \bfX, \bfW, \gamma) \exp \Big( - \sum_{i, j\in\N_i} s_{i,j} ||\bfx_i - \bfx_j||^2 \Big).
	\end{align*}
\end{small}\noindent
The second KKT condition states that the distance is preserved if $s_{i,j} \not = 0$ and $\xi_{i,j} = 0$. Let $\bfS \in \bbR^{N \times N}$ with the $(i,j)$th element as $s_{i,j}$ if $ j \in \N_i$ and $0$ otherwise. In other words, the non-zero elements of $\bfS$ stands for the distance preserving equalities.
The third KKT condition leads to $\xi_{i,j} = \frac{ s_{i,j} }{2 C}$.
Its dual problem is then obtained by substituting $\bm{\xi}$ and $q(\bfX)$ back to Lagrangian function as
\begin{small}
	\begin{align*}
	\max_{\bfS} \min_{\bfW} & - \log Z(\bfW,\bfS) - \sum_{i, j\in\N_i} s_{i,j} \phi_{i,j} - \frac{1}{4C} ||\bfS||_F^2,
	\end{align*}
\end{small}\noindent
where the partition function $Z(\bfW,\bfS)$ is further simplified by marginalizing out $\bfX$ as
\begin{small}
	\begin{align*}		
	Z(\bfW,\bfS) 
	=& \int \pi(\bfX) p(\bfY | \bfX, \bfW, \gamma)\exp \Big( - \sum_{i, j\in\N_i} s_{i,j} ||\bfx_i - \bfx_j||^2 \Big) \d \bfX \\
	=& \int \prod_{i=1}^N \N(\bfx_i | \bfzero, \bfI_d) \N( \bfy_i | \bfW \bfx_i, \gamma^{-1} \bfI_D ) \exp( - 2\Tr( \bfX^T \bfL \bfX ) ) d\bfX \\
	=& \frac{1}{\sqrt{ (2 \pi)^{N (d + D) } \gamma^{-ND} }} \exp(-\frac{\gamma}{2} ||\bfY||^2)\cdot\\
	&  \int \exp( -\frac{1}{2} \Tr( \bfX^T ( (\gamma +1) \bfI_N + 4 \bfL) \bfX) + \gamma  \Tr(\bfX^T \bfY \bfW) ) d \bfX \\
	=& \frac{1}{\sqrt{ (2 \pi)^{N (d + D) } \gamma^{-ND} }} \exp(-\frac{\gamma}{2} ||\bfY||^2) \cdot \\
	& \prod_{q=1}^d \int \exp(-\frac{1}{2} (\bfx^q)^T ( ~(\gamma +1) \bfI_N + 4 \bfL) \bfx^q + \gamma (\bfx^q)^T \bfY \bfw^q ) d \bfx^q \\
	=& \frac{1}{\sqrt{ (2 \pi)^{N D } \gamma^{-ND} }} \exp(-\frac{\gamma}{2} ||\bfY||^2) \det( (\gamma +1) \bfI_N + 4 \bfL )^{-d/2} \cdot \\ 
	& \exp ( \frac{\gamma^2}{2} \Tr(  \bfW^T \bfY^T ( (\gamma+1) \bfI_N + 4 \bfL)^{-1} \bfY \bfW )  )
	\end{align*}
\end{small}\noindent
where the third equality holds due to the orthogonal constraint, $\bfx^q$ is the $q$th column of $\bfX$, and $\bfw^q$ is the $q$th column of $\bfW$.
After substituting it back to $\log Z(\bfW, \bfS)$, we have the optimization problem.

\subsection{Proof of Proposition 2}

Taking the first derivative of the objective function with respect to $\bfZ$ and setting it to zeros, we have
\begin{small}
	\begin{align*}
	- \frac{1}{(\gamma+1)} ( \bfY - \bfZ \bfW^T ) \bfW  + \frac{4}{(\gamma+1)^2} \bfL \bfZ = 0
	\end{align*}
\end{small}\noindent
and the optimal solution $\bfZ = (\bfI_N + \frac{4}{\gamma+1} \bfL)^{-1} \bfY \bfW $. By substituting $\bfZ$ back to the objective function, we obtain
\begin{small}
	\begin{align*}
	&\frac{1}{2 (\gamma+1) } ||\bfY - \bfZ \bfW^T ||_F^2 + \frac{2}{(\gamma+1)^2} \Tr( \bfZ^T \bfL \bfZ ) \\
	=& - \frac{1}{2} \Tr(  \bfW^T \bfY^T ((\gamma+1) \bfI_N + 4 \bfL)^{-1} \bfY \bfW ) + \frac{1}{2(\gamma+1)} ||\bfY||_F^2.
	\end{align*}
\end{small}\noindent	
Since $\bfY$ is constant and independent of $\bfW$ and $\bfZ$, the proof is completed.

\subsection{Proof of Lemma 1}

Each row $\bfr_i$ of $\bfR$ in Problem (31) can be solved independently. 
%
By introducing dual variables $\alpha$, we have the Lagrangian function defined for each subproblem with respect to $\bfr_i$ as $ L(\bfr_i, \alpha) = \sum_{k=1}^K r_{i,k}  \left( ||\bfz_i - \bfc_k ||^2 + \sigma \log r_{i,k} \right) + \alpha ( \sum_{k=1}^K r_{i,k} - 1 )$.
The KKT conditions can be obtained as $ || \bfz_i - \bfc_k ||^2 + \sigma(1+\log r_{i,k}) + \alpha = 0, \forall k$ and  $\sum_{k=1}^K r_{i,k} = 1, r_{i,k} \geq 0, \forall k $, 
which lead to the following analytic solution $r_{i,k} = \exp( -{ || \bfz_i - \bfc_k ||^2 }/{\sigma}  - (1 + {\alpha}/{\sigma} ) )$. According to the KKT condition $\sum_{k=1}^K r_{i,k} = 1$, we have $\exp( 1 + {\alpha}/{\sigma} ) = \sum_{k=1}^K \exp( -{ ||\bfz_i - \bfc_k  ||^2 }/{\sigma})$. This completes the proof.
%
%
%

\subsection{Proof of Proposition 3}

According to Lemma 1, we have an analytical solution of $\bfR$ shown in (32). By substituting (32) back into the objective function of (31), we have following derivations 
\begin{small}
	\begin{align*}
	\sum_{k=1}^K \sum_{i = 1}^N r_{i,k}|| \bfz_i - \bfc_k ||^2 + \sigma \sum_{i=1}^N \sum_{k=1}^K r_{i,k} \log r_{i,k}
	= - \sigma \sum_{i = 1}^N \left\{ \log \sum_{k=1}^K \exp \left( -{ || \bfz_i - \bfc_k||^2 }/{\sigma} \right) \right\} .
	\end{align*}
\end{small}\noindent
The equality is obtained by the optimal solution  (32) and the simplex constraint of (31). The optimization problem now becomes unconstrained optimization problem 
\begin{small}
	\begin{align*}
	\min_{\bfC} - \sigma \sum_{i = 1}^N \left\{ \log \sum_{k=1}^K \exp \left( -{ || \bfz_i - \bfc_k||^2 }/{\sigma} \right) \right\} .
	\end{align*}
\end{small}\noindent
According to first order optimal condition \cite{Boyd2004}, we can obtain the optimal solution by letting the first derivative be zero, i.e., 
\begin{small}
	\begin{align*}
	\nabla_{\bfc_k} 
	&= -\sigma \sum_{i=1}^N \frac{ \exp \left( -{ || \bfz_i - \bfc_k||^2 }/{\sigma} \right) (- 2 ( \bfc_k - \bfz_i ) /\sigma )  }{ \sum_{k=1}^K \exp \left( -{ || \bfz_i - \bfc_k||^2 }/{\sigma} \right) } \\
	& = 2 \sum_{i=1}^N ( \bfc_k - \bfz_i ) \cdot \frac{ \exp \left( -{ || \bfz_i - \bfc_k||^2 }/{\sigma} \right)  }{ \sum_{k=1}^K \exp \left( -{ || \bfz_i - \bfc_k||^2 }/{\sigma} \right) } \\
	& = 2 \sum_{i=1}^N ( \bfc_k - \bfz_i ) \cdot r_{i,k} = \bfzero.
	\end{align*}
\end{small}\noindent
By solving the optimal condition problem, we have the optimal solution of $\bfc_k$ as 
\begin{small}
	\begin{align*}
	\bfc_k = \frac{ \sum_{i=1}^N r_{i,k} \bfz_i }{ \sum_{i=1}^N r_{i,k} },
	\end{align*}
\end{small}\noindent
which is equivalent to the update rule used in mean shift \cite{Cheng1995} if we consider the kernel function as a Gaussian distribution with bandwidth $\sigma$.

\subsection{Proof of Proposition 4}

The optimal solution (32) is a softmin function with respect to distance $||\bfz_i - \bfc_k||^2$. If $\sigma \rightarrow 0$, $r_{i,k} = 1$ if $k = \min_{k=1,\ldots,K} ||\bfz_i - \bfc_k||^2$, and otherwise $r_{i,k} = 0$. In the case of $\sigma \rightarrow 0$, 
we have 
\begin{small}
	\begin{align*}
	\lim_{\sigma \rightarrow 0} \sum_{k=1}^K \sum_{i = 1}^N r_{i,k}|| \bfz_i - \bfc_k ||^2 + \sigma \sum_{i=1}^N \sum_{k=1}^K r_{i,k} \log r_{i,k} 
	&= \sum_{i = 1}^N \min_{k=1,\ldots,K} ||\bfz_i - \bfc_k||^2 \\
	&= \sum_{k=1}^K \sum_{i \in \P_k}|| \bfz_i - \bfc_k ||^2,
	\end{align*}
\end{small}\noindent
where the negative entropy is equal to zero. This completes the proof.

\subsection{Proof of Lemma 2}

To prove the existence of the inverse of matrix $\frac{1+\gamma}{\gamma} (\frac{\lambda}{\gamma} \bfL + \Gamma) - \bfR^T \bfR $, we prove that this matrix is positive definite. Given any non-zero vector $\bfv \in \bbR^{K}$, we have the following derivations
\begin{small}
	\begin{align*}
	\bfv^T \left[ \frac{1+\gamma}{\gamma} (\frac{\lambda}{\gamma} \bfL + \Gamma) - \bfR^T \bfR \right] \bfv &\geq \frac{1+\gamma}{\gamma} \bfv^T \Gamma \bfv - \bfv^T \bfR^T \bfR \bfv \\
	=  &\frac{1+\gamma}{\gamma} \bfv^T \Gamma \bfv + \bfv^T ( \textrm{diag}( \bfR^T \bfR \bfone) - \bfR^T \bfR )\bfv - \bfv^T \textrm{diag}( \bfR^T \bfR \bfone)  \bfv \\
	\geq & \frac{1}{\gamma} \bfv^T \Gamma \bfv + \bfv^T \Gamma \bfv - \bfv^T \textrm{diag}( \bfR^T \bfone) \bfv \\
	= &\frac{1}{\gamma} \bfv^T \Gamma \bfv, 
	\end{align*}
\end{small}\noindent
where the first and second inequalities follow the fact that the Laplacian matrix is positive semi-definite.
If $\sum_{i=1}^N r_{i,k} > 0, \forall k$, the matrix $\frac{1+\gamma}{\gamma} (\frac{\lambda}{\gamma} \bfL + \Gamma) - \bfR^T \bfR $ is positive definite.

\subsection{Proof of Proposition 5}

With simple matrix manipulation, problem (31) with respect to $\{ \bfW, \bfZ, \bfY \}$ by fixing $\{\bfS, \bfR\}$ can be written as
\begin{small}
	\begin{align*}
	\min_{\bfW, \bfZ, \bfC} &\gamma \left[ \Tr(\bfZ \bfZ^T) - 2 \Tr( \bfR^T \bfZ \bfC^T ) + \Tr(\bfC^T \Gamma \bfC) \right]    + || \bfY - \bfZ \bfW^T ||_F^2 + \lambda \Tr(\bfC^T \bfL \bfC) : \bfW^T \bfW = \bfI_d, \nonumber 
	\end{align*}
\end{small}\noindent
where $\Gamma = \textrm{diag}( \bfone^T \bfR )$ and the Laplacian matrix over a tree represented by $\bfS$ is $\bfL = \textrm{diag}(\bfS \bfone_K) - \bfS$. Let $h(\bfW, \bfZ, \bfC)$ be the objective function of the above optimization problem. By setting the partial derivative of $h(\bfW, \bfZ, \bfC)$ with respect to $\bfC$ to zero $\partial h(\bfW, \bfZ, \bfC)/\partial {\bfC} = 2 \lambda \bfL \bfC - 2 \gamma \bfR^T \bfZ + 2 \gamma \Gamma \bfC = \bfzero$, we have an analytical solution of $\bfC$ given by
\begin{small}
	\begin{align*}
	\bfC_{\bfZ} = \left(\frac{\lambda}{\gamma} \bfL + \Gamma \right)^{-1} \bfR^T \bfZ.
	\end{align*}
\end{small}\noindent
By substituting $\bfC_{\bfZ}$ into  $h(\bfW, \bfZ, \bfC)$, we yields 
\begin{small}
	\begin{align*}	
	h(\bfW, \bfZ, \bfC_{\bfZ} ) &= || \bfY - \bfZ \bfW^T ||_F^2 + \gamma \Tr(\bfZ \bfZ^T)- \gamma \Tr\left( \bfZ^T \bfR \left(\frac{\lambda}{\gamma} \bfL + \Gamma \right)^{-1}  \bfR^T \bfZ \right).
	\end{align*}
\end{small}\noindent
Similarly, by setting the partial derivative of $h(\bfW, \bfZ, \bfC_{\bfZ})$ with respect to $\bfZ$ to zero, we obtain
\begin{small}
	\begin{align*}
	\bfZ_{\bfW} =  \left( (1+\gamma) \bfI_N - \gamma \bfR \left(\frac{\lambda}{\gamma} \bfL + \Gamma \right)^{-1}  \bfR^T \right)^{-1} \bfY \bfW.
	\end{align*}
\end{small}\noindent
According to the Woodbury formula \cite{horn2012matrix}, the optimal solution of $\bfZ_{\bfW}$ can be further reformulated as $\bfZ_{\bfW} = \bfQ \bfY \bfW$ where 
\begin{small}
	\begin{align*}
	\bfQ= & \left( (1+\gamma) \bfI_N - \gamma \bfR \left(\frac{\lambda}{\gamma} \bfL + \Gamma \right)^{-1}  \bfR^T \right)^{-1} \\
	=& \frac{1}{1+\gamma} \bfI_N - \frac{1}{(1+\gamma)^2} \bfR \left( - \frac{1}{\gamma} \left(\frac{\lambda}{\gamma} \bfL + \Gamma \right) + \frac{1}{1+\gamma} \bfR^T \bfR \right)^{-1} \bfR^T \\
	=& \frac{1}{1+\gamma} \left[ \bfI_N + \bfR \left( \frac{1+\gamma}{\gamma} \left(\frac{\lambda}{\gamma} \bfL + \Gamma \right) - \bfR^T \bfR \right)^{-1} \bfR^T  \right]\;. 
	\end{align*}
\end{small}\noindent
According to Lemma 2, the inverse matrix exists.
The objective function can thus be further written as a function with respect only to $\bfW$, which is given by $h(\bfW, \bfZ_{\bfW}, \bfC_{\bfZ_{\bfW}} ) = \Tr(\bfY^T \bfY) - \Tr( \bfW^T \bfY^T  \bfQ \bfY \bfW  )$.
The optimization problem of $h(\bfW, \bfZ_{\bfW}, \bfC_{\bfZ_{\bfW } } )$ with respect to $\bfW$ is equivalent to the following maximization optimization problem, 
\begin{small}
	\begin{align*}
	\max_{\bfW} &~ trace( \bfW^T \bfX  \bfQ \bfX^T  \bfW),\; \textrm{ s.t. }  \bfW^T \bfW = \bfI_d,
	\end{align*}
\end{small}\noindent
which is  similar to the problem of PCA and can be solved optimally by eigendecomposition on $\bfY^T \bfQ \bfY$.

\bibliographystyle{plain}
\bibliography{RBT-arxiv}

\end{document}